\documentclass[fleqn,10pt]{wlscirep}
\usepackage[utf8]{inputenc}
\usepackage[T1]{fontenc}
\usepackage{multicol}
\usepackage{setspace}
\usepackage{geometry}
\usepackage{pifont} 
\usepackage{xcolor}
\usepackage{algorithm}
\usepackage{algorithmic}
\usepackage{graphicx}

\title{Camera-based implicit mind reading by capturing higher-order semantic dynamics of human gaze within environmental context}

\author[1,2]{Mengke Song}
\author[3]{Yuge Xie}
\author[1,2]{Qi Cui}
\author[1,2]{Luming Li}
\author[1,2]{Xinyu Liu}
\author[4]{Guotao Wang}
\author[1,2,*]{Chenglizhao Chen}
\author[1,2,5]{Shanchen Pang}

\affil[1]{China University of Petroleum (East China), the Qingdao Institute of Software and the College of Computer Science and Technology, Qingdao, 266580, China}
\affil[2]{Shandong Key Laboratory of Intelligent Oil \& Gas Industrial Software, Qingdao, 266580, China}
\affil[3]{Nanjing Forestry University, School of Marxism, Nanjing, 210037, China}
\affil[4]{College of Information Science and Technology, Qingdao University of Science and Technology, Qingdao, 266061, China}
\affil[5]{State Key Laboratory of Chemical Safety, Qingdao, 266000, China}

\affil[*]{Corresponding Author: Chenglizhao Chen, cclz123@163.com}

\begin{abstract}
Emotion recognition, as a step toward mind reading, seeks to infer internal states from external cues. Most existing methods rely on explicit signals --- such as facial expressions, speech, or gestures --- that reflect only bodily responses and overlook the influence of environmental context. These cues are often voluntary, easy to mask, and insufficient for capturing deeper, implicit emotions. Physiological signal-based approaches offer more direct access to internal states but require complex sensors that compromise natural behavior and limit scalability. Gaze-based methods typically rely on static fixation analysis and fail to capture the rich, dynamic interactions between gaze and the environment, and thus cannot uncover the deep connection between emotion and implicit behavior. 
To address these limitations, we propose a novel camera-based, user-unaware emotion recognition approach that integrates gaze fixation patterns with environmental semantics and temporal dynamics. Leveraging standard HD cameras, our method unobtrusively captures users’ eye appearance and head movements in natural settings --- without the need for specialized hardware or active user participation. From these visual cues, the system estimates gaze trajectories over time and space, providing the basis for modeling the spatial, semantic, and temporal dimensions of gaze behavior. This allows us to capture the dynamic interplay between visual attention and the surrounding environment, revealing that emotions are not merely physiological responses but complex outcomes of human-environment interactions. The proposed approach enables user-unaware, real-time, and continuous emotion recognition, offering high generalizability and low deployment cost.
Experimental results demonstrate that our method improves accuracy by 13\% over traditional gaze-based techniques and exceeds physiological signal-based approaches by 2.7\% in specific scenarios, validating the powerful potential of gaze-environment interaction modeling for implicit emotion recognition in real-world applications.

\end{abstract}

\begin{document}
%\linenumbers
%\modulolinenumbers[1] % 每行加一个行号

\flushbottom
\maketitle
\thispagestyle{empty}

%\noindent Please note: Abbreviations should be introduced at the first mention in the main text – no abbreviations lists. Suggested structure of main text (not enforced) is provided below.

\begin{multicols}{2}
\setstretch{0.87}

\begin{figure*}[!t]
	\centering
	\includegraphics[width=1\textwidth]{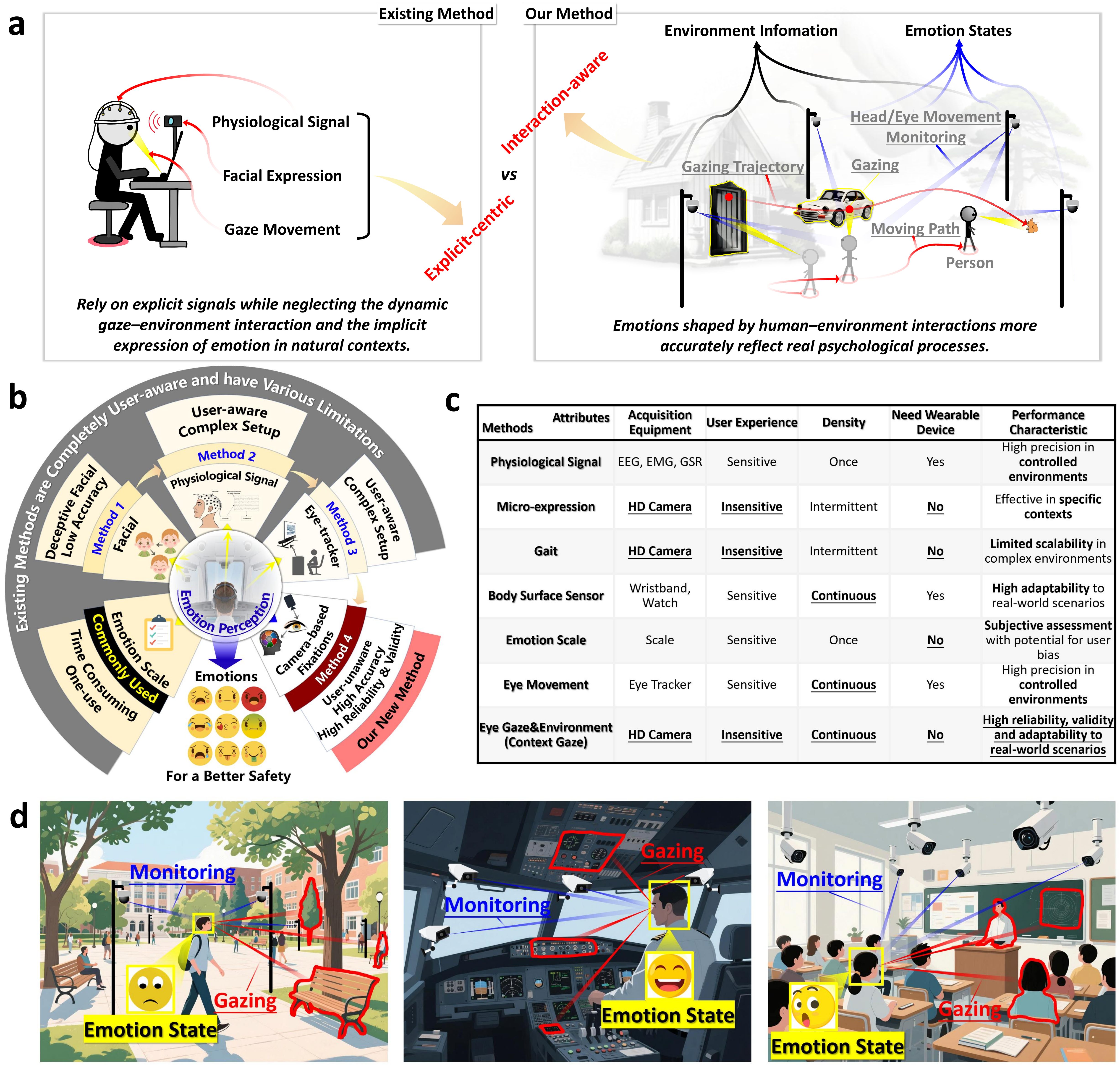}
	\vspace{-0.6cm}
	\caption{\textbf{Comparison of existing emotion recognition methods. a} Comparing traditional explicit-centric and novel interaction-aware emotion recognition methods using gaze and environmental dynamics.
		\textbf{b} Based on the depth of emotional understanding they provide and the complexity of their setup, emotion recognition methods are divided into four methods: \textbf{Method 1} (facial-based) has deceptive facial and low accuracy; \textbf{Method 2} physiological signal-based) offers deeper insights but requires complex setups and is ``user-aware''; \textbf{Method 3} (eye-tracker/gaze-based) is also complex and ``user-aware''. \textbf{Method 4} (our new method) uses camera-based fixations for high-accuracy, user-unaware recognition, providing a simple, efficient solution that overcomes the limitations of traditional methods.
		\textbf{c} Comparative Analysis of Emotion Evaluation Methods and Their Attributes. This analysis highlights the advantages of the Eye Gaze \& Environment method, which offers high accuracy and continuous data collection while minimizing user sensitivity, making it particularly effective for real-time emotion monitoring in diverse environments. ``User Experience'' describes the impact of different emotion evaluation methods on the user. ``Density'' describes the frequency of emotional data collection. 
		\textbf{d} Application scenarios of camera-based emotion state monitoring: The left image shows student monitoring on a campus, the middle image shows pilot monitoring in an airplane cockpit, and the right image shows student monitoring in a classroom. The proposed method combines user-unaware gaze tracking with environmental data for real-time, non-intrusive emotional state monitoring, with broad potential in education, transportation, healthcare, and more for personalized support and timely interventions.
	}
	%``User-unaware'' means users are not required to wear any devices and remain unaware that their data is being collected or that they are being monitored. 
	\label{fig:motivation}
	\vspace{-0.3cm}
\end{figure*}

\begin{figure*}[!t]
	\centering
	\includegraphics[width=0.95\textwidth]{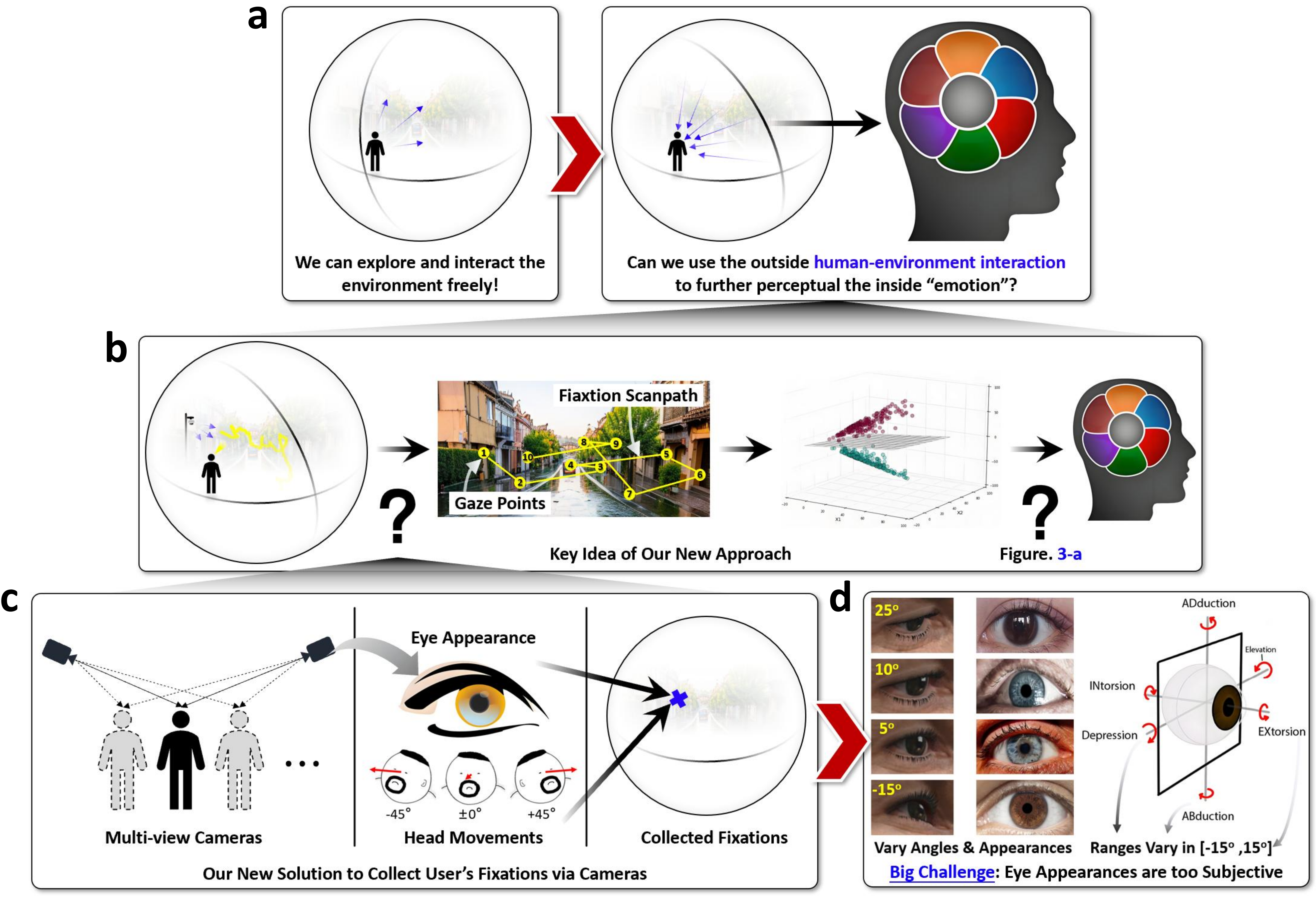}
	\vspace{-0.2cm}
	\caption{\textbf{Human-environment interaction for contextual gaze-based emotion recognition.} \textbf{(a)} illustrates the concept of leveraging human-environment interaction to infer emotions. \textbf{(b)} introduces a novel contextual gaze-based approach that combines fixation scanpaths with semantic understanding for deeper emotional insights. \textbf{(c)} A multi-camera system captures eye appearances and head movements to enable user-unaware, real-world emotion recognition. \textbf{(d)} Key challenges include variability in eye appearances and gaze angles, requiring robust gaze estimation techniques.
	}
	\label{fig:pipeline}
	\vspace{-0.3cm}
\end{figure*}

Emotion recognition is a critical step toward realizing machine-based ``mind reading'', aiming to infer internal psychological states from observable external behaviors without disrupting natural user interaction. Reading these internal emotional states --- essentially ``mind reading'' --- represents one of the most challenging yet valuable capabilities for human-computer interaction systems~\cite{MNI5}. 
Applications of emotion recognition span various fields, such as human-computer interaction~\cite{MNI3,MNI4}, healthcare, public security, education, and entertainment, offering significant benefits for user experience~\cite{zhang2020emotion}, emotional well-being~\cite{awais2020lstm,nc8,nc9,nc10}, and interpersonal communication~\cite{zhaosicheng4,zhaosicheng6,MNI2,NCt5}. However, traditional emotion recognition methods~\cite{8219396,skaramagkas2023esee,zhaosicheng5}, despite their accuracy, have limitations that have prevented the development of truly ``mind-reading'' technologies capable of understanding emotions, as shown in Figure~\ref{fig:motivation}-\textbf{a}, Figure~\ref{fig:motivation}-\textbf{b} and Figure~\ref{fig:motivation}-\textbf{c}.

At the foundation of existing emotion recognition methods are emotion scales~\cite{pekrun2017measuring,russell1989measures,nelis2011level}, considered the most commonly used due to their accuracy, as they rely on individuals self-reporting their feelings. However, these scales are limited to one-time use, are time-consuming, and unsuitable for real-time or continuous monitoring.
Building on this, facial-based methods~\cite{NCt1,lileida1,lileida2,wangshangfei1,wangshangfei2,yangjufeng2,zhaoguoying1} (Figure~\ref{fig:motivation}-\textbf{b}-\textbf{Method 1}) predominantly rely on explicit cues such as facial expressions~\cite{NCt2,NCt3,liu2023brain,xu2024class,wang2023rethinking}, voice~\cite{wagner2023dawn,martinez2024analyzing}, or bodily gestures~\cite{NCt4,lima2024st,lu2023see}. However, these signals are highly controllable, culturally variable, and easily masked, limiting their effectiveness in detecting implicit or subtle emotions in everyday settings. Although physiological signals~\cite{zhengwenming5,li2018exploring} (Figure~\ref{fig:motivation}-\textbf{b}-\textbf{Method 2}) like EEG (electroencephalography)~\cite{huang2024fbstcnet,liu2023fine,zhang2024beyond} and GSR (galvanic skin response) offer more stable emotional markers, they require wearable sensors or specialized equipment, which leads to user discomfort, high setup costs, and poor scalability in real-world environments. Gaze (eye movement)~\cite{tabbaa2021vreed,sharma2022student} (Figure~\ref{fig:motivation}-\textbf{b}-\textbf{Method 3}), as a non-contact and naturally occurring behavioral signal, has shown potential in emotion recognition. Nevertheless, conventional gaze-based methods focus mainly on static features such as fixation duration or spatial dispersion, while overlooking the dynamic interaction between visual attention and the surrounding environment. As a result, they fail to capture the underlying cognitive drivers and emotional modulations reflected in gaze behavior. This highlights the challenge of balancing simplicity, user comfort, and the depth of emotional understanding, which is essential for creating effective emotion recognition systems that are both robust and adaptable for real-world applications.

\begin{figure*}[!t]
	\centering
	\includegraphics[width=1\textwidth]{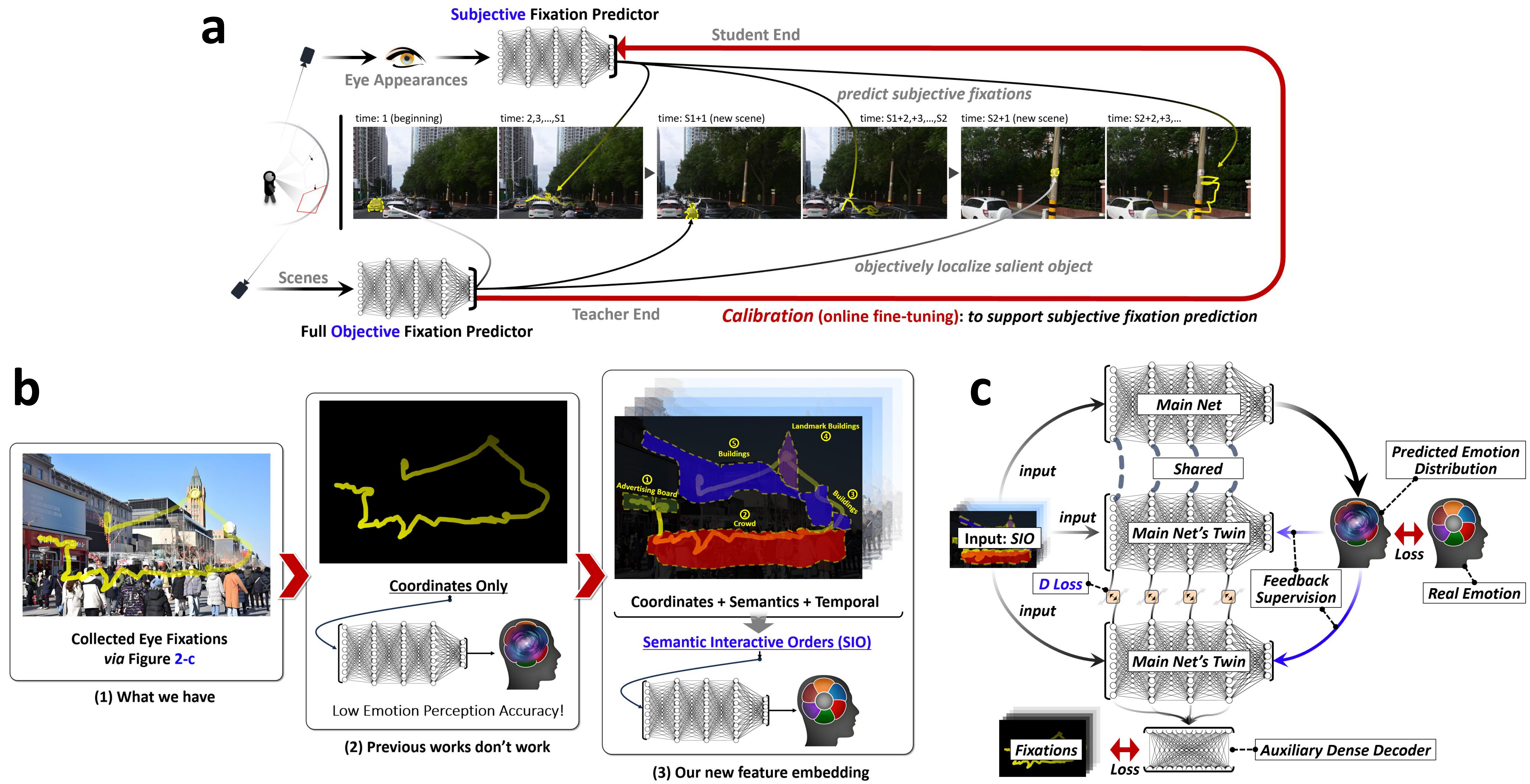}
	\vspace{-0.7cm}
	\caption{\textbf{Calibration and semantic-aware modeling for improved contextual gaze-based emotion recognition.}
			\textbf{a} shows an online calibration method that combines subjective (user-specific) and objective (scene-based) fixations to dynamically adapt gaze tracking for personalized emotion recognition.
		\textbf{b} compares traditional gaze-coordinate methods with the proposed Semantic Interactive Orders (SIO) framework, which integrates coordinates, semantics, and temporal dynamics for improved emotion detection.
		\textbf{c} showcases the proposed emotion recognition framework architecture. The model uses SIO feature embeddings as input, combined with a feedback supervision mechanism. A twin neural network structure predicts emotion distributions, and the framework employs auxiliary dense decoders and multiple loss functions to optimize performance, achieving efficient emotion recognition.}
	\label{fig:pipeline1}
	\vspace{-0.3cm}
\end{figure*}

To address these limitations, we propose a novel camera-based, user-unaware emotion recognition approach that integrates gaze fixation patterns with environmental semantics and temporal dynamics (Figure~\ref{fig:motivation}-\textbf{a}-right). Using only standard HD cameras, our method unobtrusively captures eye appearance and head movement in natural settings, without requiring any wearable devices or active participation. From these visual inputs, we estimate dynamic gaze trajectories and map them onto semantic targets in the environment, then analyze how attention to these targets evolves over time. Unlike traditional gaze methods (Figure~\ref{fig:pipeline1}-\textbf{b}(2)) that rely on static fixation statistics, our approach models how individuals dynamically interact with environmental semantic targets through gaze behavior under different emotional states (Figure~\ref{fig:pipeline}-\textbf{a, b}, Figure~\ref{fig:pipeline1}-\textbf{b}(3)). In other words, we understand emotion as an interaction between attention allocation and environmental content in complex settings, and through this ``gaze-semantics-temporal dynamics'' framework, we reveal how emotions naturally manifest in interaction with the environment.

This framework is grounded in the emotion-attention-environment interaction model, which views emotion as the outcome of both internal neural responses and externally driven cognitive regulation~\cite{lin2011influence,farshchi1999emotion}. Neuroscientific findings~\cite{kashdan2007social,richardson2004encoding} suggest that the amygdala and hippocampus jointly encode emotional content and its environmental context, with the amygdala assessing affective salience and the hippocampus forming context-based memory traces. Gaze behavior, as an overt signal of attention allocation, reflects how emotional states modulate perceptual priorities. By embedding gaze within a semantic and temporal interpretation of visual scenes, our approach captures not just what users look at, but how their attentional flow reflects emotional dynamics.

We validate our method through extensive experiments in real-world settings. The results show a 13\% improvement in emotion recognition accuracy over traditional gaze-based approaches and a 2.7\% improvement over EEG-based systems in detecting subtle emotional changes. This approach not only offers advantages in real-time processing, low cost, and privacy protection but also provides high scalability, making it applicable to various fields such as education, and public safety (Figure~\ref{fig:motivation}-\textbf{d}). Notably, our method relies solely on standard HD cameras for emotion recognition, eliminating the need for expensive specialized devices or complex sensors. The system has already been deployed in real-world environments, offering significant cost efficiency. Overall, our work provides a practical pathway toward mind-reading technologies by modeling the external behaviors associated with emotions in human-environment interactions, rather than solely relying on direct measurements of internal states, thereby achieving more accurate and unobtrusive emotion recognition.

\section*{Results}
\subsection*{A novel paradigm for emotion recognition}
%We present a novel emotion recognition framework that combines eye fixation patterns with environmental context analysis, addressing a critical gap in understanding how visual attention dynamically shapes emotional responses. Unlike traditional eye-tracking methods constrained by environmental stability requirements and limited capture ranges, our multi-camera system leverages standard HD cameras to remotely track gaze coordinates with millimeter precision, eliminating the need for specialized hardware. 
%This ``user-unaware'', camera-based approach enables continuous, unobtrusive emotion monitoring, making it both cost-effective and highly scalable for real-world applications.

Our groundbreaking contribution to the field is a new paradigm that transforms ordinary HD cameras into mind-reading devices. To address the challenges in emotion recognition, we propose a method that combines the strengths of existing approaches while overcoming their limitations. Ordinary HD cameras, which are inexpensive, easy to deploy, and widely available, could offer a promising alternative. Thus, we envision that by enabling users to interact freely with their environment, can we leverage external human-environment interactions to gain insights into internal emotions (Figure~\ref{fig:pipeline}-\textbf{a})?
To implement this, we must first identify what information an HD camera can capture from the user. While facial expressions are often unreliable for interpreting ambiguous or deceptive emotions, gaze information presents a valuable alternative. Studies have shown that gaze patterns are closely linked to emotional states, revealing indicators such as interest, stress, and cognitive engagement. This makes gaze a useful, though indirect, signal for understanding emotions.
Despite its advantages — particularly its ``user-unaware'' nature, meaning no wearable devices or active user participation are required — gaze (eye movement)-based emotion recognition has limitations. Gaze patterns primarily capture the user's line of sight and fixation behavior, but they do not directly reflect deeper emotional responses or motivations. For instance, the duration or frequency of gazing at an object may not accurately correlate with emotional intensity, as cognitive processes and external factors can influence gaze. Consequently, the relationship between gaze and emotion remains indirect, making it challenging to draw precise emotional conclusions from gaze data alone.

Building on these theoretical insights, we present a revolutionary emotion recognition framework that effectively ``reads minds'' through camera-based fixations that combines eye gaze patterns with environmental context (Figure~\ref{fig:motivation}-\textbf{b}-\textbf{Method 4}). This approach leverages the dynamic interplay between a user's visual attention and their surroundings to provide deeper insights into emotional states. 
To implement this, we developed a ``user-unaware (users are not required to wear any devices and remain unaware that their data is being collected or that they are being monitored)'' gaze tracking method that eliminates the need for specialized eye-tracking devices (Figure~\ref{fig:pipeline}-\textbf{b}). Using commonly available HD cameras, this method captures gaze points in natural, unconstrained settings and maps them onto a gaze fixation scanpath through multi-angle observations of eye appearance and head movements (Figure~\ref{fig:pipeline}-\textbf{c}). Crucially, it ensures that users remain unaware of the monitoring process, making it ideal for unobtrusive and continuous emotion monitoring. A major challenge, however, is that eye appearance is highly subjective (Figure~\ref{fig:pipeline}-\textbf{d}). Factors such as variations in gaze angles, individual differences in eye features like sclera visibility and iris size, and the complexity of 3D eye movements make it difficult to achieve consistent and accurate tracking.
Our solution to this challenge --- an online personalized calibration method --- represents a significant advance in making mind-reading technology practical and accurate in real-world environments. We incorporate this approach (see Figure~\ref{fig:pipeline1}-\textbf{a} \& Methods --- Online Personalized Calibration) that integrates subjective fixation (user-specific gaze tendencies) with objective fixation (scene-based salient points). This adaptive approach dynamically adjusts to individual differences, significantly enhancing gaze mapping accuracy and adaptability.

The core innovation of our work is the Semantic Interactive Orders (SIO) framework, which decodes the language of human gaze to reveal internal emotional states. Using the newly developed gaze tracking method (Figure~\ref{fig:pipeline}-(c)), we collect raw eye fixation points. Unlike existing approaches that rely solely on gaze coordinates, which have demonstrated low accuracy in emotion recognition (Figure~\ref{fig:motivation}-\textbf{b}-\textbf{Method 3} and Figure~\ref{fig:pipeline1}-\textbf{b}(2)), SIO combines gaze coordinates with semantic environmental information and temporal dynamics (Figure~\ref{fig:pipeline1}-\textbf{b}(3)). By mapping fixation patterns to meaningful objects and their contextual interactions over time, this framework provides a richer representation of gaze behavior, significantly enhancing the accuracy and robustness of emotion recognition.
Emphasizing a user-centric approach, the method is designed to ensure data security and anonymity, aligning with privacy regulations and addressing concerns in sensitive contexts.

\begin{figure*}[!t]
	\centering
	\includegraphics[width=1\linewidth]{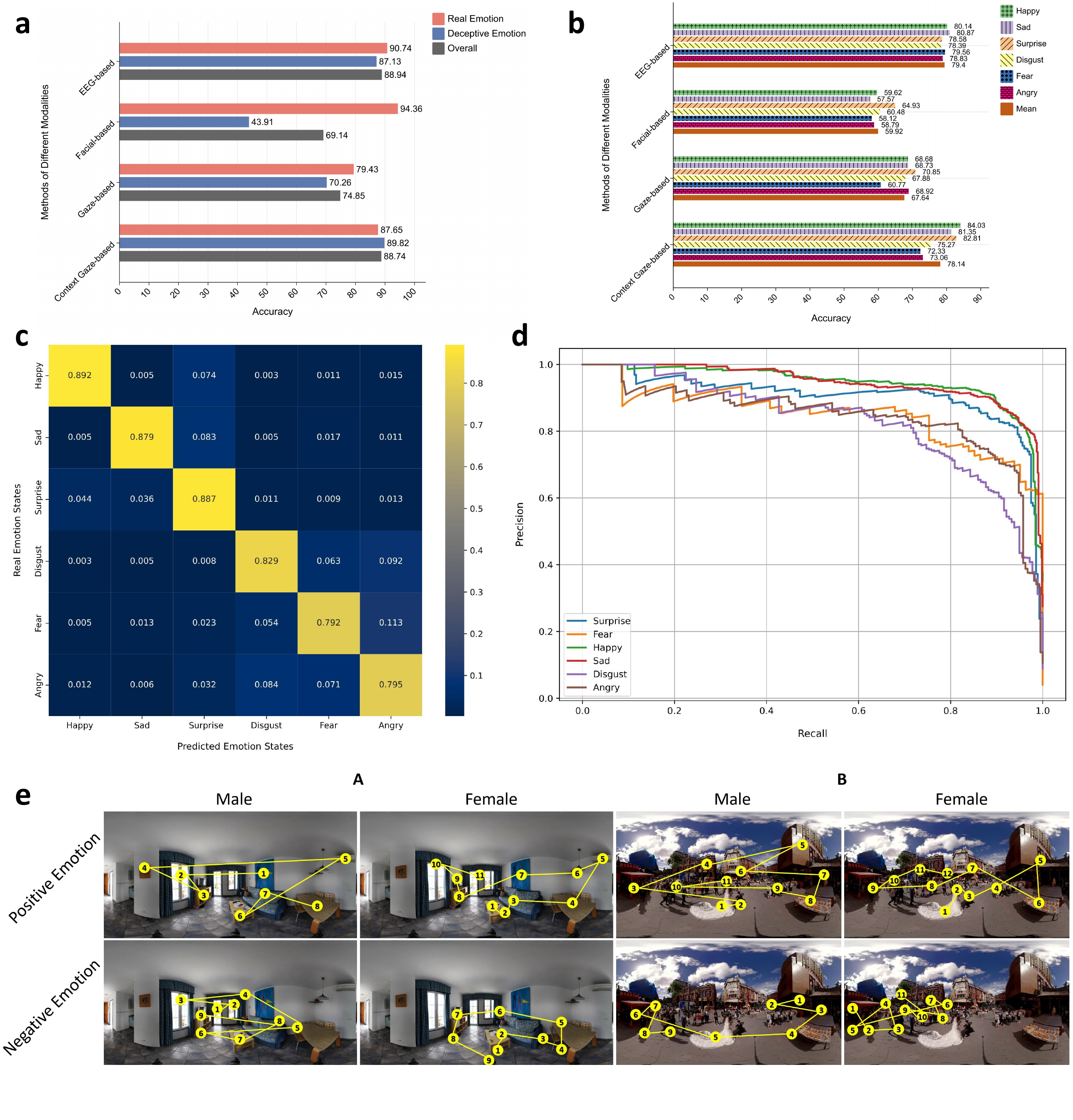}
	\vspace{-0.7cm}
	\caption{\textbf{Experimental validation of our proposed method against other methods and its performance on screen scenes.}
	Quantitative comparisons between our context gaze-based method and physiological signal (such as EEG)-based (ACTNN), facial expression-based (Toisoul), gaze (such eye movement)-based (CCER) methods. \textbf{a} In EmoGaze2D-50 dataset regarding Accuracy metric, when the users conceal their emotions, our approach can strike a balance between emotion recognition accuracy and user comfort. The facial expression-based method performs the worst. \textbf{b} In EmoGaze360-1K regarding our newly-proposed $caw\rm{F1}$ metric, our proposed emotion recognition model, EmoGazeNet, outperforms methods that rely solely on facial and eye movement analysis. Comparatively, EmoGazeNet's overall performance is only marginally less than that of EEG-based approaches, with a slight 1.26\% deficit. However, when it comes to recognizing the emotions of ``Happy'', ``Surprise'', and``Sad'', our method actually surpasses EEG techniques in terms of $caw\rm{F1}$ performance. 
	\textbf{c} The confusion matrix reflects the high accuracy of emotion state prediction, with minimal misclassification across different emotional states. \textbf{d} PR curve shows that Happy and Sad have higher precision maintained even at higher recall levels, indicating that the model performs better on these two emotions compared to others. \textbf{e} Scanpath visualization of different genders. ``Positive Emotion'': Happy, Surprise; ``Negative Emotion'': Fear, Sad, Disgust and Angry.}
	\label{fig:Screen}
	\vspace{-0.4cm}
\end{figure*}

\subsection*{Dataset Statistics for Validating the Proposed Method’s Effectiveness}
We evaluate the performance of our proposed method on three datasets. 

\textbf{Our 2D and 360-degree screen datasets.} The variability in modalities across current datasets poses challenges for direct performance comparisons. To address this, we developed a new dataset EmoGaze2D-50 inspired by the SEED-IV dataset~\cite{SEED}, which includes EEG and eye movement data, but with an expanded focus. In our dataset, we captured multimodal information --- EEG, facial expressions, eye movement data, precise fixation points, and environmental context --- as participants watched 50 videos under different emotional states. 

Since our proposed model, EmoGazeNet, processes Equirectangular Projection (ERP) images from panoramic environments, the EmoGaze2D-50 dataset, which contains only 2D images, is not suitable. Therefore, we created an another new dataset, EmoGaze360-1K, specifically for EmoGazeNet. This dataset comprises 1,000 panoramic images --- 800 indoor and 200 outdoor --- spanning 52 categories, sourced from platforms like YouTube and Vimeo. EmoGaze360-1K also includes emotional annotations for six emotional states across various modalities, including EEG, facial expressions, eye movement, precise fixation points, and environmental context. Six distinct emotions were recorded in both genuine and deceptive conditions, following strict modality-specific standards to ensure consistency and high fidelity. The dataset is split into training and testing sets with a 70/30 ratio, supporting robust ten-fold cross-validation. This dual-context approach enables fairer comparisons across emotion recognition models and offers deeper insights into how emotions are expressed when authentic or intentionally concealed (see Supplementary ``Methods - EmoGaze360-1K'' section --- EmoGaze360-1K and EmoGaze2D-50 datasets construction). Our experiments show that training on the entire EmoGaze360-1K dataset yields the best performance (Supplementary Figure 6).

\textbf{Our 360-degree real-world indoor scene dataset.} To evaluate the performance of our gaze acquisition method, we collected eye appearance data from subjects using our proposed ``user-unaware'' gaze tracking method across four distinct indoor environments, one public outdoor scene, and one driving scenario (called ``Real360''). Utilizing sophisticated post-processing techniques mentioned before, we first predict eye gaze coordinates, then transform these eye gaze coordinates into object-level regions. By analyzing the objects and corresponding environments, we can infer emotional states with greater accuracy.

\subsection*{Performance comparison between proposed deep model EmoGazeNet with existing emotion recognition methods on 2D screen dataset}
%#统一数据集对比

%\begin{figure*}[!t]
%	\centering
%	\includegraphics[width=0.5\linewidth]{../1origin_data-latest/DATA/unifiedData.pdf}
%	\vspace{-0.2cm}
%	\caption{unifiedData.}
%	\label{fig:unifiedData}
%	\vspace{-0.3cm}
%\end{figure*}

Most mainstream emotion recognition methods are trained on 2D videos where emotions are induced in participants. To ensure consistency and fairness in evaluating our proposed deep model, we used the comprehensive multimodal EmoGaze2D-50 dataset. This dataset, also based on 2D videos, incorporates a diverse array of data types, including EEG readings, facial expressions, eye movement data, precise visual fixation points, and contextual environmental information.
This dataset is particularly distinctive as it encompasses six emotional states under both deceptive and real emotional states. We conducted a comparative analysis against several state-of-the-art methods, including electroencephalography (EEG)-based method (ACTNN~\cite{ACTNN}), facial-based method (Toisoul~\cite{toisoul2021estimation}), gaze-based method (CCER~\cite{CCER}), and our context gaze-based method EmoGazeNet.

Based on the data shown in Figure~\ref{fig:Screen}-\textbf{a}, we can clearly see the significant advantage of the context gaze-based method in distinguishing between real and deceptive emotions.
The EEG-based method performed well in real emotion detection (90.74\%) with an overall accuracy of 88.94\%, though its accuracy in deceptive emotion detection dropped slightly to 87.13\%.
The facial-based method achieved the highest accuracy in real emotion detection (94.36\%), but dropped sharply to 43.91\% for deceptive emotions, highlighting its limitation in handling deceptive states.
In comparison, the gaze-based method achieved accuracies of 79.43\% and 70.26\% in real and deceptive emotion detection, respectively, with an overall accuracy of 74.85\%. It showed balanced performance but still lagged behind other methods.
Our context gaze-based method showed significant improvement over the traditional gaze-based method, with an accuracy of 89.82\% for deceptive emotions and 87.65\% for real emotions, achieving an overall accuracy of 88.74\%, close to the EEG-based method (88.94\%). Additionally, in deceptive emotion detection, our method even surpassed the EEG-based method, demonstrating its robustness and reliability across different emotional contexts. 
A two-sample t-test was conducted to compare the accuracy of our method and the traditional EEG-based method, facial-based method, Gaze-based method in deceptive emotion detection. The calculated p-value was 0.031, 0.023, 0.035 (p < 0.05), indicating a statistically significant difference, which strongly validates the superiority of our method in this aspect.

\begin{figure*}[!t]
	\centering
	\includegraphics[width=1\linewidth]{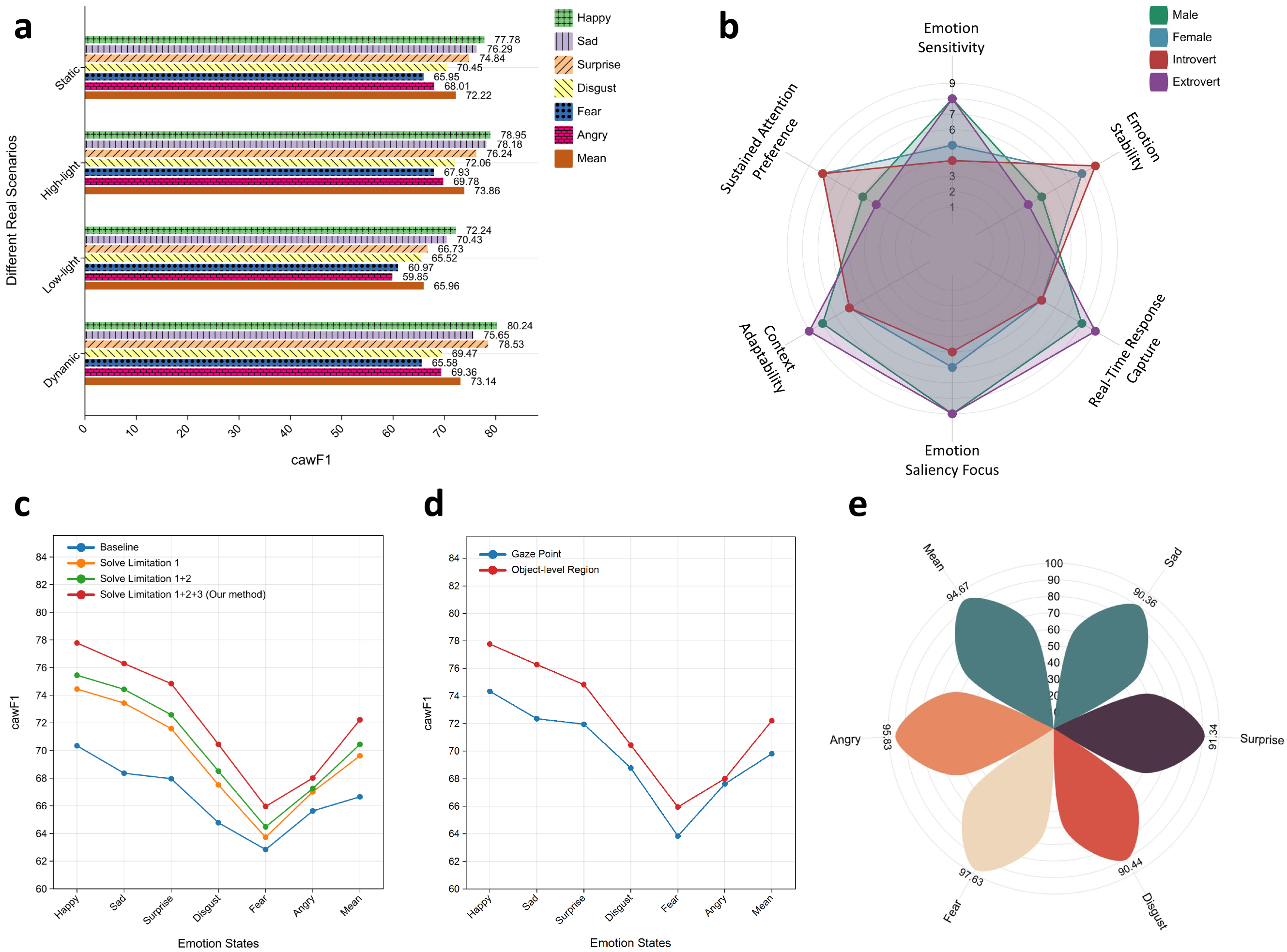}
	\vspace{-0.6cm}
	\caption{\textbf{Experimental validation of our proposed method on different settings and its performance on real scenes.} 
		\textbf{a} For real scenes (four conditions), our method achieves the best performance in high-light scene and the worst performance in low-light scene.
		\textbf{b} This radar chart illustrates that females and extroverts, with high emotional sensitivity and adaptability, are well-suited for dynamic tasks requiring quick responses, whereas males and introverts, characterized by greater emotional stability and sustained attention, are better equipped for long-term monitoring in stable environments.	
		\textbf{c} Utilizing an improved eye gaze collection method, we've resolved three key limitations, which has significantly enhanced the performance of the naive version. For the first limitation (limitation 1) --- low quality eye appearance images --- we've employed super-resolution and object-level regions to enhance clarity. To address the eye appearance variations in different users (limitation 2), we've integrated a online personality calibration process. Additionally, for the limited angles of eye appearance images (limitation 3), we developed a 3D reconstruction method to generate eye images from various perspectives.
		\textbf{d} Compared to segmenting the entire image based on gaze coordinates, mapping these coordinates to object-level regions and leveraging the sequence of these regions for emotion recognition yields better results. \textbf{e} Our research shows that using object-level regions for emotion recognition based on gaze coordinates is more effective than traditional methods. The average accuracy of correct correspondence between gaze coordinates and objects is over 94.67\%, with the ``Fear'' emotion achieving the highest accuracy at 97.63\%.}
	\label{fig:Real}
	\vspace{-0.4cm}
\end{figure*}

\subsection*{Fine-grained performance comparison between proposed deep model EmoGazeNet with existing emotion recognition methods on 360-degree screen dataset} 
We also assessed our proposed emotion recognition model EmoGazeNet on the comprehensive multimodal 360-degree screen image dataset, known as EmoGaze360-1K. We reported the $caw\rm{F1}$ performance of various emotion recognition methods --- physiological signal (EEG)-based, facial-based, gaze (eye movement)-based, and our fixation-environment integration method --- across six fine-grained emotion states.
As illustrated in Figure~\ref{fig:Screen}-\textbf{b}, ``Happy'' emotion achieved the highest $caw\rm{F1}$ score, with the context gaze-based and EEG-based methods scoring 84.03\% and 80.14\%, respectively. This high score may result from the distinct and consistent patterns associated with happiness, which makes classification easier. The ``Sad'' emotions followed with slightly lower $caw\rm{F1}$ score across the context gaze-based method, but is still higher than physiological signal (EEG)-based methods. The ``Fear'' emotion showed the lowest recognition scores, especially in the facial-based (58.12\%) and gaze (eye movement)-based (60.77\%) methods. The context gaze-based method demonstrated the second highest overall performance, with an average $caw\rm{F1}$ score of 78.14\%, surpassing both the facial-based and gaze (eye movement)-based methods, and only 1.26\% lower in overall performance compared to the EEG-based approach. This suggests that the context gaze-based method is highly effective in recognizing a range of emotional states.
A one-way ANOVA test confirmed that the differences between EmoGazeNet and physiological signal (EEG)-based method, facial-based method, gaze (eye movement)-based method were statistically significant (p = 0.015, 0.038, 0.027 < 0.05).

The confusion matrix in Figure~\ref{fig:Screen}-\textbf{c} illustrates the classification performance of our model on screen panorama data across various emotional categories. The values on the diagonal represent the model's accuracy in correctly classifying each emotion, indicating strong performance in identifying ``Happy'' (0.892), ``Sad'' (0.879), ``Surprise'' (0.887), ``Disgust'' (0.829), ``Fear'' (0.792), and ``Angry'' (0.795). ``Happy'' is primarily misclassified as ``Surprise'' (0.074), suggesting occasional confusion between these emotions, possibly due to similar facial expressions. Similarly, ``Sad'' tends to be misclassified as ``Surprise'' (0.083), reflecting challenges in distinguishing subtle emotional expressions. For the ``Surprise'' emotion, the model occasionally misclassifies it as ``Happy'' (0.044) and ``Sad'' (0.036), indicating some difficulty in distinguishing between neutral and related emotions. ``Disgust'' is more likely to be confused with ``Angry'' (0.092) and ``Fear'' (0.063), while ``Fear'' is primarily misclassified as ``Angry'' (0.113), likely due to overlapping facial cues among these emotions. Finally, ``Angry'' is mainly confused with ``Fear'' (0.071) and ``Disgust'' (0.084), which is consistent with the misclassification patterns observed for ``Fear''.

We also provided PR (Precision-Recall) curve to evaluate the performance of a model. In a PR curve, the closer the curve is to the top right corner, the better the model's performance. As shown in Fig~\ref{fig:Screen}-\textbf{d}, 
the PR curves for Fear and Disgust show high precision at low recall levels (close to 0.2 to 0.6), but precision quickly decreases as recall increases (0.6). This suggests that the model might struggle with these two emotions.
The PR curves for Surprise and Angry are more balanced, with no distinct areas of high precision, but overall, there is a good balance between precision and recall, and the curves are relatively smooth.
Happy and Sad have more prominent PR curves, with higher precision maintained even at higher recall levels, indicating that the model performs better on these two emotions compared to others.

For 360-degree indoor scenes (Figure~\ref{fig:Screen}-\textbf{e}(A)), under positive emotions, males tend to focus first on salient objects like the artwork on the blue wall, then quickly glance at windows, and finally notice the sofa and table. Females start with details on the coffee table, gradually expanding their focus to the sofa and table, and eventually to the view outside the window or door. Under negative emotions, males usually focus on windows or doors first to seek a sense of security. Then they will assess the layout of the room and pay a little attention to some bright areas. Finally, they will fix their eyes on items like sofas and coffee tables. Females begin by focusing on dark corners, then notice details on the sofa and coffee table, with their gaze primarily staying in the dark areas. 
For outdoor scenes (Figure~\ref{fig:Screen}-e(B)), under positive emotions, males first focus on the wedding scene, then quickly scan surrounding pedestrians and buildings. Females tend to first pay attention to details like the wedding dress and the expressions of the newlyweds, before expanding their focus to the entire scene, including pedestrians and buildings. Under negative emotions, males' gaze tends to be continuous, while females' line of sight is relatively shorter. Males' gaze often focuses directly on darker buildings or crowded areas in the background, and they have a shorter line of sight. Females first notice shadowy areas or details in the crowd and pay less attention to the bright wedding dress.

\subsection*{Fine-grained emotion recognition based on our eye gaze collection method in real-world indoor scene} 
We have analyzed the performance of our proposed methodology in the domain of fine-grained emotion recognition, focusing on its $caw{\rm F1}$ score across a spectrum of real-world indoor scenarios dataset Real360 that encompass diverse lighting conditions and dynamic environments. As shown in Figure~\ref{fig:Real}-\textbf{a}, ``Happy'' has the highest $caw{\rm F1}$ score in all scenarios, especially in the dynamic scenario where it reaches 80.24\%, indicating robust recognition across different conditions. In contrast, ``Angry'' and ``Fear'' show relatively lower $caw{\rm F1}$ score, with ``Angry'' in the low-light scenario being the lowest (59.85\%), suggesting that lighting conditions significantly impact the recognition of these emotions. The high-light scenario has a higher overall $caw{\rm F1}$ score (average 73.86\%), while the low-light scenario shows lower values (average 65.96\%), highlighting that sufficient lighting aids emotion recognition. Notably, the $caw{\rm F1}$ score for ``Surprise'' in the dynamic scenario (78.53\%) exceeds that in the static scenario (74.84\%), which may indicate that dynamic environments facilitate better recognition of certain emotions like ``Surprise'', suggesting that scene dynamics contribute positively to the prediction of specific emotions. A paired t-test for the comparison between high-light and low-light scenarios showed a significant difference (p = 0.042 < 0.05). We have also provided scanpaths of different genders in the four real-world scenarios under different emotion states (Supplementary Figure 7 \& 8).

Figure~\ref{fig:Real}-\textbf{c} illustrates the performance improvements of our proposed eye gaze collection method (based on the static scene) as three key limitations are progressively resolved, i.e., (1) low-quality eye appearance images, (2) eye appearance variations in different users, and (3) limited angles of eye appearance images. The baseline version (i.e., with the three key limitations), represented by the blue line, shows the lowest $caw{\rm F1}$ scores across all emotions. After addressing limitation 1 (low-quality eye appearance images), shown by the yellow line, the performance improves slightly. Further enhancements are seen with the resolution of limitation 2 (eye appearance variations in different users), indicated by the green line. Finally, resolving all three limitations, including limitation 3 (limited angles of eye appearance images), results in the highest performance across all emotions, as depicted by the red line. The improved version consistently achieves better scores, particularly for ``Happy'' and ``Sad'' emotions, with a notable improvement for ``Disgust'' and ``Angry''.

We further compared two methods for eye gaze collection: gaze point-based (blue line) and object-level region-based (red line) based on the static scene. As shown in Figure~\ref{fig:Real}-\textbf{d}, the object-level region method consistently outperforms the gaze point method across all emotions, particularly for ``Happy'', ``Sad'', and ``Surprise'' emotions, where it achieves higher $caw{\rm F1}$ scores. Both methods perform the worst on ``Fear'', with object-level regions still providing better results. Overall, object-level region mapping shows superior performance, aligning with the study's finding that accurate gaze-to-object correspondence (i.e., the gaze coordinates are precisely within the area occupied by the object) significantly improves emotion recognition, with an average correctness above 94.67\%, peaking at 97.63\% for ``Fear'' (Figure~\ref{fig:Real}-\textbf{e}). A paired-samples t-test was carried out to compare the  $caw{\rm F1}$ scores of the object-level region-based method and the gaze point-based method. The p-value was calculated to be 0.018 (p < 0.05), demonstrating a significant difference in performance between the two methods, and validating the superiority of the object-level region-based method.

\subsection*{Emotion-environment interactions across various gender and personality} 
We also explored the differences in emotion-environment interactions across various gender and personality types in Real360 dataset. We quantified the performance of participants across six indicators --- emotion sensitivity, emotion stability, real-time response capture, emotion salience focus, context adaptability, and sustained attention preference --- using a 1–10 rating scale (1 being low, 10 being high). Detailed explanations of these six indicators can be found in the Supplementary ``Methods''. The experimental results are shown in Figure~\ref{fig:Real}-\textbf{b}.

Males scored high in emotion sensitivity (8) and emotion saliency focus (9), making them responsive to emotionally rich or significant cues and well-suited for complex, dynamic environments due to high context adaptability (8). However, with lower scores in emotion stability and sustained attention preference (5), they show more frequent emotional shifts and shorter focus spans. Females, with high stability and sustained attention (8), exhibit steady emotions and prolonged focus, ideal for stable, low-dynamic contexts. Lower sensitivity and real-time response scores (5) indicate slower reactions to diverse, emotionally intense situations.

Introverts excel in emotion stability (9) and sustained attention (8), focusing well in stable, low-stimulus settings but scoring lower in sensitivity (4) and salience focus (5), making them less responsive to subtle emotional cues in dynamic environments. Extroverts score high in sensitivity (8), real-time response (9), and salience focus (9), making them quick to engage with emotional cues in varied settings. Their high adaptability (9) contrasts with lower stability and sustained attention scores (4), favoring frequent shifts in focus and making them ideal for fast-paced, interactive contexts. 
%The t-test results showed p-values of 0.002 for emotion stability, 0.003 for sustained attention, 0.0001 for sensitivity, 0.0005 for salience focus, and 0.0002 for adaptability, indicating statistically significant differences between the two groups on all six indicators.
%
%
%Independent-samples t-tests were conducted to compare the scores of males and females (average p-values<), as well as introverts and extroverts, on each of the six indicators, showed p-values < 0.05 for all comparisons, indicating statistically significant differences.

This analysis shows that males and extroverts, with their strong emotional sensitivity and adaptability, are well-suited for dynamic emotion recognition tasks, while females and introverts, with greater stability and focus, are better suited for steady, long-term monitoring in single-context environments.
These insights support security and driver monitoring. In security, males and extroverts’ heightened emotional responsiveness and adaptability aid in identifying sudden changes in high-risk individuals, while females and introverts’ stability helps in detecting abnormal behavior over longer periods. For driver monitoring, extroverts benefit from real-time alerts in complex conditions to maintain focus, whereas introverts and stable drivers, more prone to fatigue in extended sessions, can be monitored for declining attention. This approach enhances both public safety and driver security by integrating emotional and environmental interactions.

%\begin{figure*}[!t]
%	\centering
%	\includegraphics[width=1\linewidth]{../1origin_data-latest/DATA/Application.pdf}
%	\vspace{-0.7cm}
%	\caption{Field experiment to evaluate the practical application of our eye gaze collection method. \textbf{a} In the campus scenario, the context gaze-based method performed best in terms of accuracy, dynamic adaptability, and user comfort, demonstrating more stable emotion recognition compared to facial and physiological signal-based methods, while also enhancing user comfort by not requiring additional equipment. \textbf{b} In the driving simulator scenario, the context gaze-based method also showed the best performance in accuracy, F1 score, user comfort, and dynamic adaptability, while the facial-based method was significantly affected by angles and lighting in the dynamic environment, leading to the lowest performance.
%	Dynamic Adaptability: the system's ability to flexibly adjust its behavior in response to varying contextual changes; User Comfort: how easy and comfortable it is for users to interact with the system, especially without extra devices. We normalize the results of dynamic adaptability and user comfort to the range of 0-1: dynamic adaptability (high=1, medium=0.5, low=0), user comfort (high=1, medium=0.5, low=0).}
%	\label{fig:FieldExperiment}
%	\vspace{-0.4cm}
%\end{figure*}

\begin{figure*}[!t]
	\centering
	\includegraphics[width=0.9\linewidth]{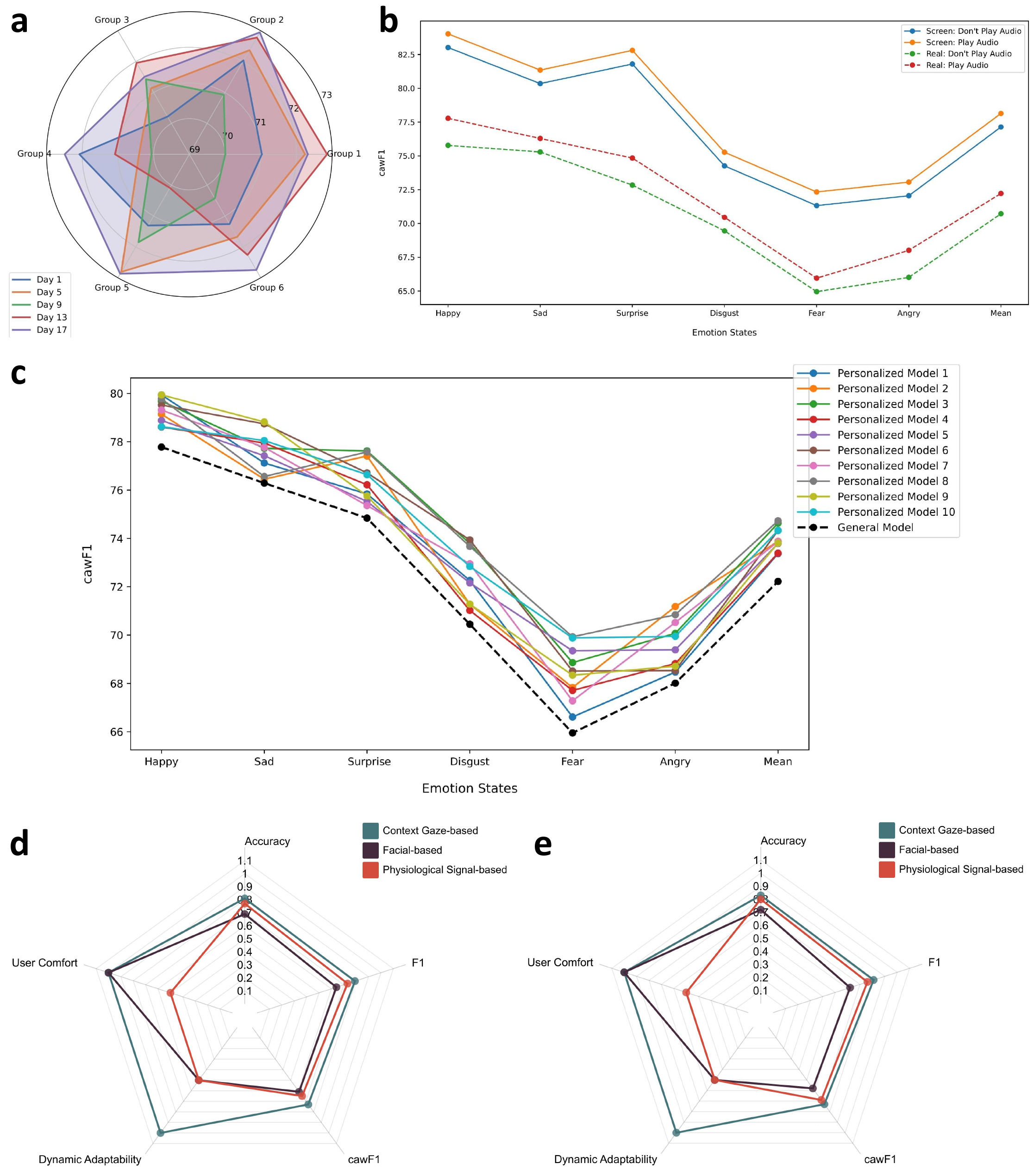}
	\vspace{-0.4cm}
	\caption{\textbf{Robustness validation of our proposed method and component evaluation.} \textbf{a} Long-term stability monitoring experiment indicates that during the long-term monitoring period, the proposed method can continuously provide a relatively consistent level of accuracy in emotional recognition. \textbf{b} Emotional enhancement experiment shows that playing corresponding emotional audio on emotion recognition when collecting gaze points in real scenes and when viewing 360-degree images on a screen are efficient to promote emotion recognition accuracy. \textbf{c} Personalized models have higher average accuracies than the general model, indicating that personalized training can enhance the performance of emotion recognition. \textbf{d} and \textbf{e} are field experiments to evaluate the practical application of our eye gaze collection method. \textbf{d} In the campus scenario, the context-aware gaze method outperformed facial and physiological approaches in accuracy, dynamic adaptability, and user comfort, offering more stable emotion recognition without requiring additional equipment. \textbf{e} In the driving simulator, it again achieved the highest accuracy, F1 score, adaptability, and comfort, while facial methods suffered from angle and lighting issues, resulting in the lowest performance. Dynamic adaptability refers to the system’s responsiveness to changing contexts; user comfort reflects ease of use without extra devices. Both metrics are normalized to [0–1]: high=1, medium=0.5, low=0.}
	\label{fig:RobustnessComp}
	\vspace{-0.4cm}
\end{figure*} 

\subsection*{Field experiment to evaluate the practical application of our eye gaze collection method} 
To evaluate the practical application of the proposed eye gaze collection method, we designed two field experiment scenarios: a campus environment and a driving simulator environment. These scenarios, with their unique environmental characteristics and emotional demands, provide a comprehensive assessment of the method’s robustness and effectiveness in real-world settings.

\textbf{Campus Scenario Experiment Setup.}
The experiment was conducted in an open campus area (e.g., a campus square) to simulate a dynamic and varied social and natural environment. Fifteen university students (6 females, 9 males, aged 22-28) participated and were asked to simulate six different emotional states (such as Happy, Angry, and Surprise) induced by videos or images. During the experiment, eight high-definition cameras captured participants' eye appearance from different angles, which was then combined with gaze-environment interactions for emotion recognition. Additionally, participants wore Apple Watches to monitor skin conductance response (GSR) and had their facial expressions recorded.

Results in Figure~\ref{fig:RobustnessComp}-\textbf{d} showed that in the campus scenario, our context gaze-based method, which incorporates gaze-environment interactions, outperformed the other methods, achieving an accuracy of 81.45\%, F1 score of 0.81, and $caw{\rm F1}$ score of 0.73. This was significantly higher than the facial-based method (71.42\%, 0.62, and 0.62, respectively) and the physiological signal-based method (79.62\%, 0.74, and 0.65). One-way ANOVA was performed to compare the accuracy of the three methods. The p-value was 0.025 (p < 0.05), indicating significant differences among the methods. Post-hoc tests showed that the differences between our context gaze-based method and the facial-based method, as well as the physiological signal-based method, were statistically significant (p < 0.05 for both comparisons). Our context gaze-based method also scored highest in dynamic adaptability and user comfort (both 1), demonstrating its suitability for complex, open environments without requiring additional wearable equipment. In contrast, the facial-based and physiological signal-based methods showed lower dynamic adaptability and physiological signal-based methods showed lower user comfort, with facial expression recognition especially affected by lighting and angle, leading to less stable results.

\textbf{Driving Simulator Scenario Experiment Setup.}
To simulate driving conditions and collect relevant data, we used a driving simulator that presented various traffic scenarios (e.g., emergency braking, traffic congestion). Ten drivers (7 males, 3 females, aged 20-29) participated and were tasked with handling different driving challenges and emotional stimuli. The experiment used three screens to simulate a realistic driving view, displaying front, left, and right window perspectives to replicate real driving conditions. Only one high-definition camera was used in front of the simulator to capture eye appearance, and facial expressions and GSR data were recorded simultaneously.

In the driving scenario, as shown in Figure~\ref{fig:RobustnessComp}-\textbf{e}, our context gaze-based method also outperformed the other methods, with an accuracy of 83.85\%, F1 score of 0.81, and $caw{\rm F1}$ score of 0.73. By comparison, the facial-based method achieved lower scores (72.62\%, 0.62, and 0.58), while the physiological signal-based method scored slightly higher than the facial-based method (82.15\%, 0.79, and 0.71). A one-way ANOVA was performed to compare the accuracy of the three methods. The calculated p-value was 0.032 (p < 0.05), indicating significant differences among the methods. Post-hoc tests further revealed that the differences between our context gaze-based method and the facial-based method were significant with a p-value of 0.021 (p < 0.05), and the difference between our method and the physiological signal-based method also reached statistical significance with a p-value of 0.045 (p < 0.05). In terms of dynamic adaptability and user comfort, our context gaze-based method scored the highest (both 1). While the facial-based method had high user comfort (1), it showed lower adaptability (0.5), indicating it was more affected by head movements and changes in lighting within the vehicle. The physiological signal-based method scored low in both dynamic adaptability and user comfort (0.5).

Overall, our context gaze-based method incorporating gaze-environment interactions demonstrated higher accuracy, F1 scores, and adaptability in both the campus and driving simulator scenarios, enabling more accurate emotion recognition in complex environments. Additionally, it provided superior user comfort by avoiding the need for additional wearable devices. These results suggest that the gaze-based method is particularly advantageous for emotion recognition tasks requiring high adaptability and user comfort.

\textbf{Future Research Directions in Mental Health and Security.} The positive results from our experiments suggest several promising future research areas. First, psychological state assessment could benefit from our method, enabling continuous, user-unaware monitoring of emotional shifts in real-time. This would be particularly valuable in environments like campus settings, where emotional fluctuations in students could be tracked seamlessly as they move through various activities (Figure~\ref{fig:motivation}-\textbf{d}-left). Second, early psychological disorder screening could be facilitated by identifying subtle emotional cues that indicate mental health issues such as anxiety or depression, even before they become fully apparent. This application could extend to sensitive settings, such as monitoring pilots in an airplane cockpit, where early detection of emotional distress is critical (Figure~\ref{fig:motivation}-\textbf{d}-middle). Third, our method could enhance smart classroom environments by monitoring students' emotional states in real-time, providing valuable insights into their engagement, stress levels, and overall well-being (Figure~\ref{fig:motivation}-\textbf{d}-right). These research directions underscore the broad potential of gaze-based emotion recognition in advancing mental health monitoring and enhancing public security.

\subsection*{Robustness evaluation}
% 长期稳定性监测实验
% 情绪加强实验
% 个性化模型训练实验

To prove the robustness of the proposed method, we conducted three experiments.

We conducted a long-term stability monitoring experiment designed to track the emotional recognition results of the same group of users over different time periods, e.g., from day to day in a real-world static scenario, where emotional recognition tests are conducted every six hours, twice a day, over a span of multiple days to sufficiently capture the daily changes in the user's emotional state. Specifically, the experiment design involves testing with two people per group, which may help reduce the impact of individual differences on the results. As shown in Figure~\ref{fig:RobustnessComp}-\textbf{a}, the data shows the fluctuation in $caw{\rm F1}$ scores. The data displays the $caw{\rm F1}$ score of emotional recognition from Day 1 to Day 17.
The $caw{\rm F1}$ score fluctuate around the 70\% to 73\% mark, indicating that the proposed method maintains relatively stable performance over the long-term monitoring period. Despite daily fluctuations, there is no significant overall decline or upward trend in $caw{\rm F1}$ scores, suggesting that the proposed method has good long-term stability. On Day 9 in Group 1, the $caw{\rm F1}$ score reached its lowest point at 70.01\%, while on Day 13, it reached its highest point at 72.86\%.
These peaks and troughs may be related to various factors, such as changes in the user's state, environmental factors, or minor changes in test conditions.
In consecutive tests, the $caw{\rm F1}$ score showed only minor changes. From the 1st to 5th day in Group 3, it rose slightly from 70.22\% to 71.14\%. From the 13th to 17th day, it dropped marginally from 71.96\% to 71.51\%, indicating overall stability.
Such short-term fluctuations may be due to random errors or minor changes in the user's emotional state. Looking at the overall data from Day 1 to Day 17, the $caw{\rm F1}$ score seems to fluctuate around a central value of approximately 71.5\% to 72\%.
This indicates that during the long-term monitoring period, the proposed method can continuously provide a relatively consistent level of accuracy in emotional recognition.

We implemented emotional enhancement experiment aimed to compare the impact of playing or not playing corresponding emotional audio on emotion recognition when collecting gaze points in real static scenes and when viewing 360-degree images on a screen. As shown in Figure~\ref{fig:RobustnessComp}-\textbf{b}, 
for viewing 360 images on a screen, without playing emotional audio, the average $caw{\rm F1}$ score is 77.14\%; with playing emotional audio, the average $caw{\rm F1}$ score increases to 78.14\%. 
For collecting gaze points in real scenes, without playing emotional audio, the average $caw{\rm F1}$ score is 70.72\%; with playing emotional audio, the average $caw{\rm F1}$ score increases to 72.22\%. 
%In both scenarios, playing emotional audio improved the average $caw{\rm F1}$ score of emotion recognition. For viewing 360 images on a screen, the accuracy increased by 1.29\%; in real scenes, the accuracy increased by 1.26%.
%For specific emotions, such as ``Happy'' and ``Sad'', the $caw{\rm F1}$ score of recognition improved when emotional audio was played, indicating that emotional audio may enhance the visual recognition of these emotions.
Playing emotional audio had a certain enhancing effect on the recognition $caw{\rm F1}$ score of most of emotions, although the extent of improvement varied.

We conducted personalized model training experiment aimed to train an individualized emotion recognition model for each user and assess the effectiveness of personalized models in enhancing emotion recognition for individual users based on the static scene in real world. The experiment also compared the performance differences between personalized and general models. Data was collected every two hours, seven times a day, for the training of the personalized model. As shown in Figure~\ref{fig:RobustnessComp}-\textbf{c}, personalized models show varying average $caw{\rm F1}$ scores across different users (numbered 1 to 10), roughly ranging from 73.37\% to 74.73\%. The general model has an average $caw{\rm F1}$ scores of 72.22\%, serving as the baseline for comparison with the performance of personalized models. In most cases, personalized models have higher average $caw{\rm F1}$ score than the general model, indicating that personalized training can enhance the performance of emotion recognition.

\begin{figure*}[!t]
	\centering
	\includegraphics[width=1\linewidth]{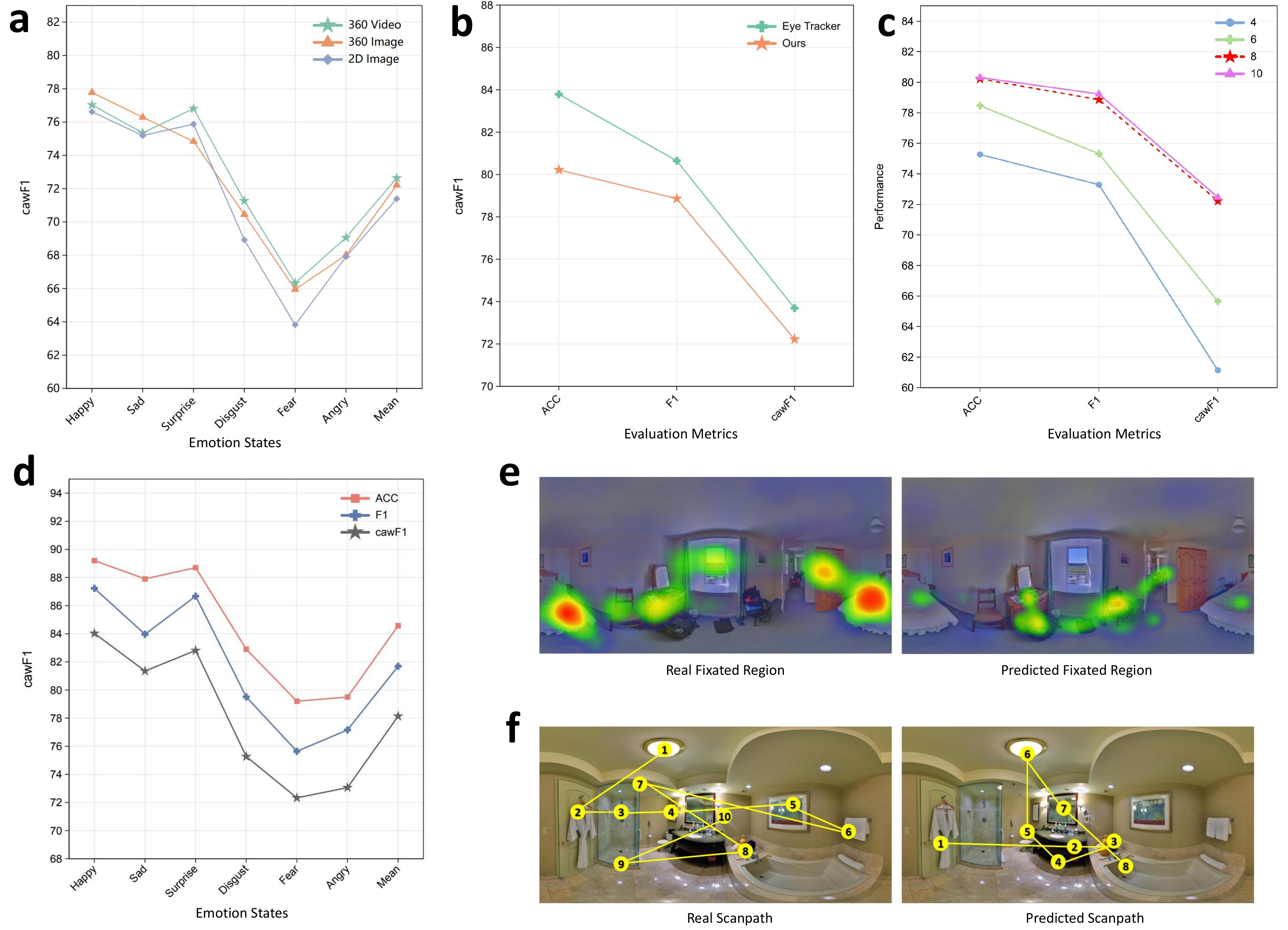}
	\vspace{-0.7cm}
	\caption{\textbf{Ablation studies on Real360 dataset.} \textbf{a} When collecting users' gaze points in the real-world scene of the Real360 dataset, while 360-degree videos generate better user responses than 360-degree images, the overall differences between 2D images, 360-degree images, and 360-degree videos are minimal. \textbf{b} While eye trackers offer a more precise way to capture user's gaze points, our proposed method stands out for its exceptional effectiveness in monitoring emotions in real-time across a variety of settings. \textbf{c} The performance is highest when using 8 HD cameras for eye appearance acquisition. \textbf{d,e,f} are experimental validation of our proposed evaluation metric $caw{\rm F1}$ on EmoGaze360-1K dataset. \textbf{d} Quantitative illustration of our proposed method on EmoGaze360-1K dataset regarding Accuracy (ACC), F1 and $caw{\rm F1}$ metrics. The rigor of the new metric is reflected in the fact that it requires the model to not only recognize emotions, but also accurately determine which areas of the picture participants are focusing on based on their emotional states. The low score of the new metric reflects a gap in the model's fine-grained understanding of emotions. \textbf{e} In the ``Fear'' state, the participant's gaze may have been more focused on the safe bed, and the model did not capture this particular area of attention well. The new metric $caw{\rm F1}$ therefore scored low. \textbf{f} The model could correctly predict emotional states but failed to capture the differences in gaze scanpath, thus, the new metric $caw{\rm F1}$ scored low.}
	\label{fig:Ablation}
%	\vspace{-0.1cm}
\end{figure*}

\subsection*{Component evaluation and proposal evaluation metric} 
%We compared the effects of different types of visual stimuli on emotion recognition by examining 2D images, 360-degree panoramic images, and 360-degree videos. As shown in Figure~\ref{fig:Ablation}-a, 360-degree panoramic images demonstrated the highest $caw{\rm F1}$ score for recognizing ``Fear'' (71.21\%), ``Happy'' (81.26\%), and ``Sad'' (79.89\%), suggesting that 360-degree panoramic images may be more effective in capturing the distinct features of these emotions. For ``Neutral'' emotion, however, 2D images performed better (82.81\%) compared to 360-degree panoramic images (81.87\%), indicating that 2D images might be more adept at capturing the subtleties of neutral expressions. When comparing 360-degree videos and panoramic images, the average $caw{\rm F1}$ score difference was minimal (76.89\% vs. 76.58\%), with slight variations across emotions, such as a 0.39\% higher $caw{\rm F1}$ score for ``Angry'' emotion in panoramic images. Overall, while 2D images generally showed slightly better performance in emotion recognition, the differences across data types were relatively small, particularly between 360-degree videos and images.

We compared the impact of different visual stimuli on emotion recognition by examining 2D images, 360-degree panoramic images, and 360-degree panoramic videos on Real360 dataset As shown in Figure~\ref{fig:Ablation}-\textbf{a}, 360 videos demonstrated superior performance across most emotional dimensions. Specifically, 360 videos achieved the highest recognition rates in “Surprise” (76.81\%), “Disgust” (71.27\%), “Fear” (66.33\%), and “Angry” (69.06\%), and also attained the highest overall mean score of 72.64\%.
On the other hand, 360 images excelled in conveying “Happy” (77.78\%) and “Sad” (76.29\%) emotions, outperforming both 360 videos and 2D images. This indicates that 360 images have a distinct advantage in eliciting both positive and negative emotions.
In contrast, 2D images showed slightly lower scores across all emotional dimensions, with “Happy” at 76.62\% and “Sad” at 75.19\%. However, they still maintained relatively high recognition rates in “Surprise” (75.87\%) and an overall mean of 71.39\%.

%In order to verify the validity of the proposed method, 360 images of real scenes were displayed on a computer screen, and the gaze coordinates were collected by eye tracker and compared with those collected by the proposed method. As shown in Figure~\ref{fig:Ablation}-b, the eye tracker has an accuracy of 83.79\%, while the method proposed in this paper has an accuracy of 80.22\%. This indicates that the eye tracker is slightly more accurate in collecting gaze coordinates than the method proposed in this paper. The F1 score for the eye tracker is 80.65\%, and for the method proposed in this paper, it is 78.86\%. The F1 score is the harmonic mean of precision and recall, and a higher F1 score implies better overall performance. On this metric, the eye tracker also shows a slightly better performance. The $caw{\rm F1}$ score for the eye tracker is 73.69\%, while for the method proposed in this paper, it is 72.22\%. The eye tracker shows higher performance across all three evaluation metrics compared to the method proposed in this paper. This may indicate that the eye tracker, as a mature tool for gaze data collection, has certain advantages in accurately capturing gaze points. However, although the eye tracker's performance is slightly higher on these metrics, the performance of the method proposed in this paper is also quite close, which may indicate that the method is practical, especially in situations where resources are limited or rapid deployment is required.

To validate our method, we compared gaze coordinates collected by our approach and an eye tracker using 360 images on Real360 dataset. As shown in Figure~\ref{fig:Ablation}-\textbf{b}, the eye tracker showed slightly higher accuracy (83.79\%) compared to our method (80.22\%), with an F1 score of 80.65\% for the tracker and 78.86\% for our approach. The eye tracker also had a higher $caw{\rm F1}$ score (73.69\% v.s. 72.22\%). These results suggest that while the eye tracker is more accurate in collecting gaze data, our method performs comparably and is practical for resource-limited or rapid deployment scenarios.

Figure~\ref{fig:Ablation}-\textbf{c} shows the performance variations when using different number of HD cameras during eye appearance acquisition. The use of 8 cameras appears to strike an optimal balance between capturing comprehensive eye appearance data and managing the computational complexity and potential data redundancy. Eight cameras provide sufficient coverage of the eye region to capture the necessary details for accurate gaze prediction without overwhelming the system with excessive data. This balance helps maintain high accuracy and F1 scores, as the model can efficiently process the captured data without being hindered by unnecessary information.

We also evaluated each component of the proposed emotion recognition model EmoGazeNet (see Supplementary Table 1).
The evaluation matrix clearly demonstrates that the system reaches its peak performance in terms of ACC (80.22\%), F1 Score (78.86\%), and $caw{\rm F1}$ (72.22\%) when all key components, such as Scanpath-guided Region Generation, Primary Classification Branch, Auxiliary Classification Branch, and Scanpath-guided Classification Branch, are engaged, highlighting the synergistic impact of these elements on overall system performance. We also compared different scanpath prediction methods and different choices of base encoder in the Generator (Supplementary Figure 2 \& 3).

We conducted extensive experiments on EmoGaze360-1K dataset to demonstrate that the proposed metric, $caw{\rm F1}$, is more effective than existing metrics like accuracy and F1 score. As shown in Figure~\ref{fig:Ablation}-\textbf{d}, the quantitative results reveal that $caw{\rm F1}$ scores are generally lower than the other two metrics. For instance, in the ``Fear'' emotional state, participants' gaze tended to focus more on the safe bed, a specific area the model failed to capture accurately, leading to a lower $caw{\rm F1}$ score (Figure~\ref{fig:Ablation}-\textbf{e}). Additionally, when recording participants' gaze scanpaths while they viewed complex scene images under different emotional states, the model successfully predicted the emotional states but struggled to account for the variations in gaze scanpaths. Consequently, $caw{\rm F1}$ scored lower, highlighting its stricter evaluation metric (Figure~\ref{fig:Ablation}-\textbf{f}).

\section*{Discussion}
This study introduces a novel framework for emotion recognition that deeply integrates gaze behavior with environmental semantics and temporal dynamics, enabling continuous and non-intrusive monitoring of internal emotional states. By moving beyond the reliance on explicit cues such as facial expressions, speech, and gestures, our approach leverages the wealth of information embedded in natural gaze shifts and attention allocation during everyday human-environment interactions. Extensive empirical evidence demonstrates strong generalizability and robustness across a wide range of scenarios (see Fig.\ref{fig:Screen}, Fig.\ref{fig:Real}), laying a solid technical foundation for emotion sensing and decoding in real-world contexts.

At the theoretical level, our work resonates with recent advances in neuroscience concerning the interplay among emotion, attention, and environmental context. The results reveal that the generation and regulation of emotion are not isolated internal processes but are fundamentally embedded in the continuous interaction between individuals and their environment (see Fig.\ref{fig:motivation}, Fig.\ref{fig:pipeline}, Fig.\ref{fig:pipeline1}). By tracking spatial transitions in gaze, semantic targets of attention, and their temporal evolution, we show that emotional states can manifest in subtle and complex patterns of perception and behavior. These findings provide not only new theoretical insights for affective computing but also empirical evidence for understanding the functional mechanisms of emotion in social interaction and cognitive regulation (see Supplementary Fig. 7 and Supplementary Fig. 8).

We constructed multimodal datasets in a variety of settings and systematically compared our environment- and gaze-based approach with traditional methods relying on facial cues, gaze points, or physiological signals. The results show clear advantages in both accuracy and applicability, especially in challenging environments involving emotional concealment or complex attentional shifts (see Fig.\ref{fig:Screen}-\textbf{a,b} and Fig.\ref{fig:Real}-\textbf{a}). Notably, modeling attention at the level of semantic objects allows the system to capture the natural flow of emotional states in real-world scenarios, moving beyond static gaze coordinates or single perceptual features. This is crucial for enabling emotion recognition systems to operate effectively in complex, real-life applications.

Individual differences also play a significant role in emotion-environment interaction (see Fig.\ref{fig:Real}-\textbf{b}). Our analyses indicate that factors such as gender and personality significantly influence gaze patterns, emotional sensitivity, and stability, providing a solid empirical basis for the development of more refined and personalized emotion recognition systems (see Supplementary ``Methods''). Furthermore, our long-term and dynamic environment experiments confirm the stability and adaptability of the method (see Fig.\ref{fig:RobustnessComp}-\textbf{a}), suggesting broad applicability in domains such as education, public safety, and intelligent driving (see Fig.~\ref{fig:RobustnessComp}-\textbf{d,e}).

Despite these advances, several limitations remain. Under conditions of extreme lighting, occlusion, or rapid head movement, the accuracy of gaze estimation and emotion recognition still needs improvement. Moreover, the complexity of emotion --- shaped by culture, age, and psychological factors --- calls for larger and more diverse datasets to enhance model generalizability and fairness. From an ethical and privacy standpoint, although our method is non-intrusive and low-observability by design, large-scale deployment requires rigorous standards for data collection and usage to safeguard user consent and data security.

In summary, this work not only extends the theoretical and methodological foundations of emotion recognition but also provides a new path for interdisciplinary research and the development of emotional intelligence systems in real-world scenarios. As foundational theories and technologies continue to evolve, emotion recognition based on dynamic modeling of environment, attention, and emotion is poised to have far-reaching impact in areas such as mental health monitoring, human-computer interaction, intelligent education, and public safety.

\section*{Methods}
\subsection*{Camera-based gaze tracking method}
\textbf{Collection setting.}
We strategically position eight HD cameras around a designated area, ensuring full coverage and eliminating blind spots. This setup allows individuals to move freely within the space while continuously capturing images from all angles (see Figure~\ref{fig:pipeline1}-a). For each position within the space, a segment of the panoramic image corresponding to the individual's field of view (FOV) is projected onto a 2D plane. Detailedly, we use a third-person multi-camera panoramic modeling approach to ensure a ``user-unaware'' solution by generating a panoramic model of the scene from any location, capturing the user's gaze interaction with the environment (see Methods --- Third-person multi-camera panoramic modeling section).

To achieve this, we recruited 30 annotators (18 females and 12 males, aged 18 to 28) to collect data on their eye appearance and regions of focus. Prior to data collection, participants underwent emotion induction through video and image stimuli. Each participant collected data six times in the same scene, under six distinct emotional states. Based on Paul Ekman’s basic emotion theory~\cite{ekman1971constants}, we also categorize emotions into six types, i.e., ``Angry'', ``Disgust'', ``Fear'', ``Happy'', ``Sad'', and ``Surprised''. To avoid memory residual interference (i.e., carryover effects from one emotion affecting the regions of focus in subsequent emotions), there was a two-day interval between each data collection session for different emotions within the same scene. In total, we collected data from one static indoor scenes, two high/low light indoor scene, and one dynamic outdoor scenes.

\textbf{Gaze mapping.}
By analyzing the visual appearance of the eyes in this 2D projection, our system predicts the coordinates of the gaze point using existing advanced gaze prediction algorithms (see Supplementary Figure 4). This projection and prediction process occurs at short intervals, resulting in a comprehensive dataset of gaze coordinates mapped onto the 360-degree panoramic image over time. Notably, our findings indicate that a projection interval of 0.1 seconds optimizes the accuracy of gaze point collection (see Supplementary Figure 5). Since the current advanced gaze prediction algorithms learn the mapping between eye appearance and coordinates directly, ignoring the subjectivity of gaze data and the variability in eye appearance across individuals. This results in reduced generalization and accuracy of the mapping. To improve, we then employ an online personalized calibration method to reduce the interference caused by individual differences in eye appearance using our collected data (see Figure~\ref{fig:pipeline1}-\textbf{a} and Methods section). Further, due to variations in camera angles, lighting, and other factors, collected eye images may lack clarity, so we enhance them with super-resolution. Additionally, most existing datasets for gaze estimation include only limited head angles, restricting the range of eye appearances. Our method provides greater flexibility in gaze angles, capturing diverse viewpoints. To bridge this gap, we perform 3D reconstruction on the facial data from the existing gaze prediction dataset (e.g., ShanghaiTechGaze~\cite{lian2018multiview}), generating eye appearance data from various angles. This data then retrains the model to better fit our requirements. The details of collection and post-processing of eye appearance data in real-world environments can be seen in Supplementary ``Methods --- Gaze Point Collection Process'' section and Supplementary Algorithm 1 and Algorithm 2.

This innovative approach allows for precise and continuous tracking of eye gaze across a wide area, overcoming the limitations and discomfort associated with traditional eye tracking devices. The collected gaze coordinates are then integrated with the 360-degree panoramic images and fed into our proposed emotion recognition model, EmoGazeNet, enabling robust and accurate emotion detection. 
Furthre, to assess the accuracy of gaze point collection accuracy, we have introduced an object-box-based evaluation metric: if the gaze coordinates fall within the object box, it is considered accurate; otherwise, it is deemed to have a significant error. This metric provides a straightforward way to evaluate the precision of gaze estimation systems, ensuring that the predicted gaze points are closely aligned with the actual areas of interest within the environment (see Methods section).

\subsection*{Context Gaze-based deep model}
Directly mapping eye appearance, gaze coordinates, and environmental context can lead the model to learn superficial visual patterns, associating environment with emotional states without understanding deeper gaze-related correlations. To address this, we introduce EmoGazeNet, a novel GAN-based model designed to capture meaningful, context-aware interactions between gaze behavior, emotional states, and environmental cues.
EmoGazeNet takes two primary inputs: (1) a panoramic environment represented by an ERP (Equirectangular Projection) image, and (2) sequential gaze coordinates reflecting human-environment interactions. These gaze coordinates form a scanpath used to segment the ERP image into distinct object patches, ordered according to the sequence of viewing. Different arrangements of these patches effectively reflect variations in emotional states.

As shown in Figure~\ref{fig:pipeline1}-\textbf{c} \& Supplementary Figure 1, EmoGazeNet, based on Generative Adversarial Networks (GANs), is designed with two main components: the Generator (Main Net) and the Discriminator (Main Net's Twins). The Generator is responsible for generating the probability distribution of emotion categories, while the Discriminator's task is to distinguish between real and generated data. Through adversarial training, both components continuously improve, with the Generator becoming better at generating realistic emotional state predictions, and the Discriminator sharpening its ability to differentiate between true and synthesized data. This enables the model to not only learn the direct relationships between gaze and emotional states but also refine its understanding of the deeper, context-aware correlations between eye movements and the surrounding environment. Details of EmoGazeNet are shown in the Supplementary ``Methods - EmoGazeNet model achitecture''.

\subsection*{Online personalized calibration}
Gaze data is inherently subjective, as individuals exhibit widely varying gaze patterns in identical situations. For example, when observing the same artwork, some may focus on the main character, while others may be drawn to background details or color contrasts. These variations stem from personal interests, preferences, and observation habits, complicating the adaptability and generalizability of emotion recognition models. Furthermore, the mapping between eye appearance and gaze coordinates varies significantly among individuals, making it challenging for traditional regression models to achieve high accuracy in gaze tracking. Given that gaze is a fine-grained external expression, even small tracking errors can significantly interfere with emotion detection, emphasizing the need for high precision.

To address these challenges, we propose an online personalized calibration method that integrates subjective fixation (user-specific gaze tendencies) and objective fixation (scene-based salient points) to enhance gaze mapping accuracy and adaptability (Figure~\ref{fig:pipeline}-\textbf{c} and Figure~\ref{fig:pipeline1}-\textbf{a}). This method focuses on leveraging two key factors influencing gaze behavior: head motion and gaze state transitions. First, when the head is stationary, head movement data has minimal influence on gaze accuracy. However, during head movement initiation or cessation, visual inertia causes the gaze to align roughly with the head’s direction, offering a valuable reference point. A multi-camera system captures images from multiple angles, and head pose estimation algorithms calculate pitch, yaw, and roll. By monitoring changes in head angles, the system identifies movement start and stop points. At these moments, a “strong hint” mechanism provides an initial gaze range, reducing errors caused by individual differences.
Second, gaze transitions between two states: ``scanning'' and ``fixation''. In the scanning state, the gaze moves rapidly over a wide area, while in the fixation state, it focuses on a specific object. Distinguishing these states enables more precise gaze tracking. An initial gaze mapping model, combined with head pose data, estimates the approximate gaze position. For static objects, saliency detection identifies the most prominent object as the gaze coordinate, aligning with the objective fixation. For dynamic objects, motion detection techniques like optical flow pinpoint the movement’s starting point as the precise gaze coordinate. This integration of head motion and gaze states ensures robust, individualized gaze tracking across diverse scenarios.

Building on this foundation, the calibration process adapts dynamically to user behavior and environmental changes, starting with a global initialization and continuing through ongoing fine-tuning.

At system initialization (timestamp t=1), a global calibration process aligns subjective and objective fixation (Figure~\ref{fig:pipeline1}-\textbf{a}). During this phase, the system collects eye appearance data (e.g., pupil shape, gaze direction) and head movement data (pitch, yaw, roll) using a multi-camera setup. This data forms the basis for aligning the subjective and objective gaze references. To achieve this, a teacher-student model framework is used, inspired by knowledge distillation. 
The teacher model analyzes the scene to identify salient objects, such as static targets (e.g., cars, trees) or dynamic movement starting points, establishing an objective fixation reference. The student model, which is personalized to the user, predicts gaze points based on subjective fixation tendencies and compares them with the teacher's outputs. This comparison serves to refine the student model through continuous learning, gradually adjusting it to better align with both the scene's characteristics and the user's preferences, enhancing the system's adaptability and accuracy over time.

During significant head movements or scene transitions (e.g., yaw or pitch exceeding thresholds at timestamps t=S1+1 and t=S2+1), dynamic calibration is triggered to adjust gaze predictions. Visual inertia temporarily aligns gaze with head direction, allowing the ``strong hint'' mechanism to narrow the gaze range. Simultaneously, the teacher model updates salient object detection, particularly for dynamic regions, and the student model is fine-tuned by integrating motion starting points and salient targets. This process distinguishes between scanning and fixation states, offering broad gaze ranges during scanning and precise targets during fixation.

Finally, during scene transitions, the system performs online fine-tuning within the first 200–300 milliseconds --- a critical window when visual attention is primarily driven by objective saliency rather than cognitive or emotional factors~\cite{awh2012top,theeuwes2010top}. The system reacquires eye appearance data and head movement updates, while the teacher model reanalyzes salient objects in the new scene. This enables rapid calibration of the student model to reflect new scene characteristics. By aligning subjective fixation with the most prominent static or dynamic features (objective fixation), the system ensures precise gaze tracking even in complex and dynamic environments.

\subsection*{Third-person multi-camera panoramic modeling}
Gaze often exhibits distinct ``first-person'' characteristics~\footnote{``First-person'' characteristics refer to the unique perspective where the user's gaze directly aligns with their view of the environment, making it challenging to capture and interpret objectively.} in its interaction with the environment. To represent this perspective, conventional methods typically rely on wearable devices, such as head-mounted cameras. However, while these devices can effectively capture the user's field of view, they also increase the user's burden and reduce the overall user experience.
To solve this challenge, we propose a ``third-person multi-camera panoramic modeling'' approach, using a multi-camera approach from a third-person perspective to generate a panoramic model of the scene from any location, ensuring a ``user-unaware'' solution (see Figure~\ref{fig:pipeline}-\textbf{c} \& Supplementary ``Methods --- Third-Person Multi-Camera Panoramic Modeling''). 

We adopt a problem decomposition strategy, breaking down the complex panoramic sphere generation task into three subproblems: static background fine reconstruction, local foreground object appearance generation, and foreground-background high-realism fusion. Since the static background is relatively stable, we use computationally expensive methods (such as ReconFusion) to reconstruct the base background of the panoramic sphere. For dynamic foreground objects, a ``lightweight'' approach is needed for high-quality generation and fusion. The specific solution consists of two parts:

First, the foreground semantic skeleton captures key information about movable objects in the scene, such as spatial coordinates, size, appearance, and semantics, using multiple complementary camera views. Due to the differences in object representation across various viewpoints, we need to achieve ``common alignment'' and ``differential complement'' of object-level information in a lightweight manner through a ``weakly supervised'' model. Specifically, a subspace clustering approach is used to establish initial mappings of foreground objects from different camera angles, and through self-iteration, the local structure matching is optimized to create a ``sparsely structured'' and ``semantically rich'' foreground semantic skeleton. This method significantly reduces the complexity of panoramic sphere generation and meets real-time requirements.

Second, to reduce panoramic sphere generation's computational overhead, we simplify processing by utilizing pre-reconstructed backgrounds and foreground semantic skeletons. Our approach generates target object appearances from desired viewpoints using semantic skeletons, then fuses them with backgrounds. Camera parameters and object poses from the semantic skeleton enable efficient local-to-global fusion with enhanced realism. We further optimize through ``weight-sharing, alternating training'', using a single model for both viewpoint generation and fusion, improving quality without additional computational costs.

The advantage of this method lies in problem decomposition, ensuring high-quality generation while reducing the demand for computational resources. By using multi-camera joint generation to create high-quality ``first-person perspective'' panoramic spheres, we can represent the interaction between the viewpoint and the environment in a ``user-unaware'' manner, laying a crucial foundation for subsequent research.

\subsection*{Object-box-based evaluation metric for gaze point collection accuracy}
To evaluate the accuracy of the collected gaze coordinates, we propose a method based on the object's bounding box. This method assumes that every object in the scene is labeled with a bounding box, and the model's predicted gaze coordinates should fall within the bounding box of an object. We judge the accuracy of the prediction based on whether the gaze coordinates fall within the bounding box. If the coordinates fall inside the object's bounding box, the prediction is considered accurate; if they fall outside the bounding box, the prediction is considered to have a large error.

To describe this process specifically, we assume that the model's predicted gaze coordinates are $(x_p,y_p)$, while the bounding box of the closest object is represented by the coordinates of its top-left and bottom-right corners, $(x_{min},y_{min})$ and $(x_{max},y_{max})$, respectively.

First, we check if the gaze coordinates satisfy the following conditions to confirm whether they fall inside the object's bounding box:
\begin{equation}
	x_{min} \le x_{p} \le x_{max}\ \ \ \text{and}\ \ \ y_{min} \le y_{p} \le y_{max},
\end{equation}
if these conditions hold, the gaze coordinates $(x_p,y_p)$ are within the object's bounding box, and the prediction is considered accurate.

Second, if the gaze coordinates do not satisfy the above conditions, i.e.:
\begin{equation}
	x_{p}<x_{\min }\ \ \ \text{or}\ \ \ x_{p}>x_{\max }\ \ \ \text{or}\ \ \ y_{p}<y_{\min }\ \ \ \text{or}\ \ \ y_{p}>y_{\max },
\end{equation}
then the gaze coordinates $(x_p,y_p)$ are outside the object's bounding box, and the prediction is considered to have a large error.

Third, we can define an accuracy evaluation function Accuracy, which takes a value of 1 (accurate) or 0 (inaccurate), using the following formula:
\begin{equation}
	\textbf{A}=\left\{\begin{array}{ll}
		1, & \text { if }\  x_{\min } \leq x_{p} \leq x_{\max }\ \ \ \text{and}\ \ \ y_{\min } \leq y_{p} \leq y_{\max }, \\[0.5em]
		0, & \text { otherwise }.
	\end{array}\right.
\end{equation}

Here, $\textbf{A}=1$ indicates that the prediction is accurate, meaning the gaze coordinates fall within the object's bounding box; $\textbf{A}=0$ indicates that the prediction is inaccurate, meaning the gaze coordinates are outside the bounding box.

\subsection*{Implementation Details}
The EmoGazeNet model is developed and implemented using PyTorch in Python with CUDA. Model training is performed on an NVIDIA Geforce RTX 3090 graphics processing unit (GPU).
We use the Adam optimizer with the learning rate of 0.001 to train the EmoGazeNet model for 1000 epochs with batch size of 16. The complement training process takes around 17 hours. The model has 50 GFLOPs and 17.84 million parameters. 

To evaluate the performance of EmoGazeNet, we report three key metrics: Accuracy (Acc), F1 score, and Contextual Attention Weighted F1 Score ($caw{\rm F1}$). Acc measures the overall correctness of the model's predictions, calculated as the ratio of correct predictions (both true positives and true negatives) to the total number of predictions made.
F1 score provides a balanced measure of precision and recall, which is particularly useful in scenarios with imbalanced class distributions.
Contextual Attention Weighted F1 Score ($caw{\rm F1}$) is our proposed evaluation metric tailored specifically for emotion recognition tasks. This metric not only assesses classification performance but also incorporates fixation-context consistency. By doing so, $caw{\rm F1}$ evaluates the model's ability to correctly classify emotions while simultaneously understanding the relationship between eye fixations and the environmental context, thereby providing a more comprehensive assessment of the model's performance.

\subsection*{Proposed evaluation metric}
Traditional multi-classification evaluation metrics such as precision, recall, and F1 score are usually used for basic performance measurement, but they may not be sufficient to capture the complexity of emotion recognition tasks, particularly when considering the interaction between emotional states and visual attention. These conventional metrics focus solely on the accuracy of emotion classification without taking into account the specific areas of the environment that individuals might focus on under different emotional states. As a result, models evaluated using these metrics might appear to perform well, even when they fail to accurately predict the gaze patterns or fixation points that are crucial for understanding the emotional context.

Targeting at this issue, we propose a comprehensive evaluation metric that addresses these limitations by integrating both classification performance and fixation-context consistency into a single evaluation metric --- Contextual Attention Weighted F1 Score ($caw{\rm F1}$). 
Unlike traditional metrics, $caw{\rm F1}$ not only assesses the model's ability to correctly classify emotions but also evaluates how well the model can predict the areas of the environment that are most relevant to the observed emotional state. This makes the metric more rigorous and reflective of the model's true understanding of the interplay between emotion and attention. By incorporating gaze patterns into the evaluation, $caw{\rm F1}$ ensures that models are held to a higher standard, where successful emotion recognition is closely tied to accurate environmental context interpretation. The metric can be defined as:
\begin{equation}
	caw{\rm F1}=\frac{\sum_{i=1}^{n}{\rm FCC}_i\cdot b{\rm F1}_i}{\sum_{i=1}^{n}{\rm FCC}_i} ,
\end{equation}
where $n$ is the number of samples, $b{\rm F1}_i$ is the balanced F1 score for the $i$-th sample. ${\rm FCC}_i$ is the fixation-context consistency score for the $i$-th sample, used to measure the consistency of the model between the detected viewpoints and the context of the environment. ${\rm FCC}$ can be calculated by:
\begin{equation}
	{\rm FCC}=\frac{1}{n}\sum_{i=1}^{n}(\alpha\cdot {\rm Sim}(v_i^{local},e_i^{local})+\beta\cdot {\rm Sim}(v_i^{global},e_i^{global})),
\end{equation}
where $n$ is the number of samples, $v_i^{local}$ and $v_i^{global}$ are the local and global fixation feature vectors of the $i$-th sample, $e_i^{local}$ and $e_i^{global}$ is the local and global environment context feature vectors of the $i$-th sample. $\alpha$ and $\beta$ are the weight parameters which satisfy $\alpha+\beta=1$. ${\rm Sim}$ is used to compute the cosine similarity between fixation features and environmental context features.

The features of fixation and environment context regarding local and global conditions can be extracted by pre-trained convolutional neural networks (e.g., ResNet, VGG, etc.)
For local features $v_i^{local}$ and $e_i^{local}$, we extract features within a certain area around the gaze point. For example, features within a fixed-size window around the point of gaze and corresponding environmental information and may be extracted.
For global features $v_i^{global}$ and $e_i^{global}$, we extract global features from the entire image to capture overall fixations and environmental information.

\subsection*{Principles for emotion category selection}
In selecting emotion categories, we followed principles of theoretical representativeness and experimental feasibility. Six basic emotions --- ``Angry'', ``Disgust'', ``Fear'', ``Happy'', ``Sad'', and ``Surprised'' --- were chosen based on Ekman's theory, covering the core spectrum of human affect. These emotions are easily elicited and annotated in controlled settings, supporting reliable multimodal data collection and enabling direct comparison with previous studies. This choice also ensures compatibility and generalizability within existing affective computing and psychological research frameworks.

\subsection*{Privacy Protection and Ethical Considerations}
When applying gaze-based emotion recognition methods, privacy protection is crucial. Although this approach uses a ``user-unaware'' monitoring system, it is important to ensure users' privacy is not compromised. To address this, we anonymize all collected emotional data, ensuring it cannot be traced to specific individuals. Additionally, data is encrypted during storage and transmission, and sensitive information is anonymized to remove personal identifiers, enhancing security. We also comply with relevant laws and regulations to safeguard user privacy. Future work will further explore and refine privacy protection mechanisms, ensuring ethical application in fields like public safety and mental health.

%\section*{Code availability}
%The source codes, datasets and results are publically available and can be downloaded from \url{https://github.com/songsook/EmoGaze360}. 
%Additional requests or inquiries about the code can be made to songsook@163.com.

%Topical subheadings are allowed. Authors must ensure that their Methods section includes adequate experimental and characterization data necessary for others in the field to reproduce their work.

\bibliography{refs}

%\noindent LaTeX formats citations and references automatically using the bibliography records in your .bib file, which you can edit via the project menu. Use the cite command for an inline citation, e.g.  \cite{Hao:gidmaps:2014}.

%For data citations of datasets uploaded to e.g. \emph{figshare}, please use the \verb|howpublished| option in the bib entry to specify the platform and the link, as in the \verb|Hao:gidmaps:2014| example in the sample bibliography file.

%\section*{Acknowledgements (not compulsory)}
%
%Acknowledgements should be brief, and should not include thanks to anonymous referees and editors, or effusive comments. Grant or contribution numbers may be acknowledged.

%\section*{Author contributions}
%M.S. and C.C. conceived the project. M.S. designed and implemented the deep learning model. Q.C., A.H., H.Q., and DP.F. analyzed the data. M.S., Q.C., and S.P. wrote the manuscript. All authors read and revised the manuscript.

%Must include all authors, identified by initials, for example:
%A.A. conceived the experiment(s),  A.A. and B.A. conducted the experiment(s), C.A. and D.A. analysed the results.  All authors reviewed the manuscript. 

\section*{Competing interests}
The authors declare no competing interests.

%\section*{Additional information}
%
%
%\textbf{Correspondence} and requests for materials should be addressed to Chenglizhao Chen.

%To include, in this order: \textbf{Accession codes} (where applicable); \textbf{Competing interests} (mandatory statement). 
%
%The corresponding author is responsible for submitting a \href{http://www.nature.com/srep/policies/index.html#competing}{competing interests statement} on behalf of all authors of the paper. This statement must be included in the submitted article file.

%Figures and tables can be referenced in LaTeX using the ref command, e.g. Figure \ref{fig:stream} and Table \ref{tab:example}.

\end{multicols}

\clearpage
\section*{Supplementary Methods}

\subsection*{Gaze Point Collection Process}
The whole processing procedure is as shown in Algorithm~\ref{algorithm1} and Algorithm~\ref{algorithm2}, which consists of the following steps.

\subsubsection*{Camera Placement and Calibration}
To comprehensively cover all possible human viewpoints within the room, eight cameras are placed in various corners to ensure that every corner is within the field of view, minimizing the potential for viewpoint loss. A chessboard calibration method is used to calibrate these eight cameras, obtaining the internal and external parameters of each camera (such as focal length, distortion coefficients, position, and orientation) through computer vision techniques. This establishes an accurate camera coordinate system and field of view, providing a foundation for subsequent positioning and viewpoint prediction.

\subsubsection*{Human Body Positioning and Viewpoint Range Determination}
Using multi-view image reconstruction technology and triangulation, the human body position within the room is determined from multiple perspectives. By combining the internal and external parameters of the cameras and matching key points from different views, the triangulation method is used to calculate the 3D coordinates of the human body. This positioning process is crucial for subsequent viewpoint prediction, ensuring the accuracy of the human body’s position to infer the viewpoint range.

\subsubsection*{Third-Person Multi-Camera Panoramic Modeling}
The goal is to create a panoramic model of the surrounding environment centered around the human. A problem decomposition strategy is used to break the complex task of generating a panoramic sphere into three parts: static background reconstruction, local foreground generation, and foreground-background fusion. First, a high-precision method (such as ReconFusion) is used to reconstruct the relatively stable static background, establishing the base environment. For dynamic foreground objects, lightweight processing is used, extracting key features from multiple camera views through foreground semantic skeleton extraction. By combining weakly supervised models, "common alignment" and "differential supplement" are realized to establish the semantic skeleton of foreground objects from different perspectives, optimizing local structure matching and ensuring the realism of dynamic objects. Finally, through the "weight sharing and alternating training" method, high-efficiency generation and fusion interactions are achieved while reducing computational costs, successfully generating panoramic sphere images with high realism.

\subsubsection*{Human Eye FOV Acquisition and 2D Image Projection}
Facial recognition or head pose estimation algorithms are used to obtain the face direction (such as pitch, yaw, and roll angles), determining the direction of the eye gaze. The user’s forward direction is set as the center of the FOV area, with a predefined FOV angle (e.g., 120° or 130°) covering the eye's attention area. The panoramic image is mapped onto a sphere, and the specified FOV area is extracted from the sphere and projected onto a 2D plane to form the user's FOV view. FOV images captured by multiple cameras can be stitched and aligned through multi-view fusion techniques to improve the accuracy and clarity of the view, dynamically adjusting the user's head direction and gaze to update the FOV area in real-time.

\subsubsection*{Online Personalized Calibration and Collection of Viewpoint Coordinates}
Head movement and gaze physiological characteristics are combined with images captured by multiple cameras and head pose detection algorithms to accurately capture head direction. When head movement starts or ends, the gaze is roughly consistent with the head direction, providing auxiliary information for viewpoint coordinates. The "online strong prompt" mechanism is used to adjust the coordinate range during the start and end of movements, improving the accuracy of viewpoint prediction and reducing errors due to individual differences.

\subsubsection*{Eye Super-Resolution and Viewpoint Coordinate Prediction}
Since the human field of view is limited and a single capture cannot cover all viewpoint coordinates, viewpoint coordinates need to be predicted at short intervals and projected onto the overall 2D panoramic image. The collection frequency of all cameras is synchronized to ensure data consistency and accuracy, and eye super-resolution technology is used to enhance clarity and improve the accuracy of predictions.

\subsubsection*{Fine Viewpoint Prediction}
Viewpoint coordinate prediction can utilize existing viewpoint prediction models such as ShanghaiTechGaze. Since the data collection angles of this models are limited, the facial data needs to be reconstructed in 3D to generate eye images from different angles, expanding the freedom of perspective and retraining the model to adapt to more viewpoint angles. Moreover, eye perspective changes do not affect viewpoint coordinates; the same viewpoint coordinates may correspond to different eye appearances from multiple perspectives.

\subsubsection*{Coordinate Space Consistency}
To ensure that the 360 images we collect and project and the viewpoint data collected by ShanghaiTechGaze are analyzed in the same coordinate space, the coordinates need to be remapped so that data from different sources is processed within a unified coordinate space.

\subsubsection*{Accuracy Improvement: Object Area Detection}
Viewpoint coordinates are used as signal input, combined with object detection algorithms, to detect the object area where the viewpoint coordinates are located. By converting relatively coarse viewpoint predictions into precise object-level detection, this compensates for viewpoint errors caused by camera distance. This approach not only improves positioning accuracy but also precisely identifies the object of gaze, which is beneficial for emotional recognition and environmental interaction applications.

\subsubsection*{Viewpoint Coordinate Prediction Accuracy Evaluation}
To evaluate the accuracy of viewpoint coordinate predictions, an evaluation criterion is set: if the predicted coordinates fall outside the object box, the prediction is considered to have a significant error; if they fall within the box, it is considered relatively accurate, based on the principle that eye gaze focuses on the object area. The closer the predicted result is to the object area, the more accurate it is considered to be.

\subsubsection*{Multiple Trials and Optimization}
Multiple trials are conducted to verify the stability and accuracy of the algorithm, adjusting camera positions and numbers to maximize coverage. Continuous optimization is carried out to gradually improve the accuracy of viewpoint predictions, ultimately achieving stable and high-precision viewpoint coordinate predictions.

\begin{algorithm} 
	\caption{Gaze Point Collection and Prediction Process - Part 1} 
	\begin{algorithmic}[1] 
		\STATE \textbf{Step 1: Camera Placement and Calibration} 
		\FOR{each camera $C_i$ where $i = 1, \dots, 8$} 
		\STATE Place $C_i$ at designated corner to maximize room coverage, ensuring each corner is within view to minimize gaze point loss.
		\STATE Perform calibration using a checkerboard pattern to obtain intrinsic parameters (focal length $f_i$, distortion coefficient $k_i$) and extrinsic parameters (position $\mathbf{t}_i$ and orientation $\mathbf{R}_i$).
		\STATE Calculate camera matrix $K_i = \begin{bmatrix} f_{x_i} & 0 & c_{x_i} \\ 0 & f_{y_i} & c_{y_i} \\ 0 & 0 & 1 \end{bmatrix}$ for $C_i$, and save $K_i, \mathbf{R}_i, \mathbf{t}_i$. 
		\ENDFOR
		
		\STATE \textbf{Step 2: Human Position and Gaze Range Estimation}
		\STATE Capture image set $\{I_i\}$ from each $C_i$.
		\STATE Using multi-view geometry, apply triangulation on key points $\{\mathbf{p}_i\}$ across views to compute the 3D coordinates $\mathbf{P}$ of the human position.
		\STATE Compute $\mathbf{P}$ as:
		\[
		\mathbf{P} = \operatorname{triangulate}(\{\mathbf{p}_i\}, \{\mathbf{R}_i\}, \{\mathbf{t}_i\})
		\]
		\STATE Store $\mathbf{P}$ as the initial reference point for subsequent gaze range estimation.
		
		\STATE \textbf{Step 3: Third-Person Panoramic Modeling (Centered on Human Position $\mathbf{P}$)}
		\STATE Divide the modeling task into three parts: background reconstruction, foreground generation, and fusion.
		\STATE \textbf{1. Background Reconstruction:} Use high-precision algorithms (e.g., ReconFusion) to reconstruct the static background around $\mathbf{P}$, resulting in the background map $B$.
		\STATE \textbf{2. Foreground Generation:} For each dynamic object $F_j$ around $\mathbf{P}$, extract semantic skeleton $\mathbf{S}_j$ using multi-view analysis.
		\STATE \textbf{3. Foreground-Background Fusion:} Integrate $B$ and $\{\mathbf{S}_j\}$ using weight-sharing and alternate training methods to reduce computational load. Finalize the panoramic model $\Pi$ as:
		\[
		\Pi = \operatorname{fuse}(B, \{\mathbf{S}_j\})
		\]
		
		\STATE \textbf{Step 4: Field of View (FOV) Extraction and 2D Projection}
		\STATE Use head orientation angles $(\theta, \phi, \psi)$ (pitch, yaw, roll) to determine gaze direction $\mathbf{g}$.
		\STATE Set FOV angle $\alpha$ (e.g., 120°), with gaze direction $\mathbf{g}$ as the central vector.
		\STATE Map $\Pi$ to a spherical coordinate system, then extract the FOV region $\Pi_{\alpha}$ centered at $\mathbf{g}$.
		\STATE Project $\Pi_{\alpha}$ to a 2D plane for display as $\Pi_{\alpha}^{2D}$.
	\end{algorithmic}
	\label{algorithm1}
\end{algorithm}

\newpage

\begin{algorithm} 
	\caption{Gaze Point Collection and Prediction Process - Part 2} 
	\begin{algorithmic}[1] 
		\STATE \textbf{Step 5: Personalized Gaze Calibration and Data Collection} 
		\STATE For detected head movements $\mathbf{h}$, adjust FOV center $\mathbf{g}$ based on the ``strong online hint'' mechanism, ensuring precise gaze alignment. 
		\STATE Store adjusted gaze coordinates as $\mathbf{g}_{calib}$.
		
		\STATE \textbf{Step 6: Gaze Super-Resolution and Coordinate Prediction}
		\STATE Synchronize camera capture frequency $\nu$ for consistent data.
		\STATE Apply super-resolution processing on each frame to enhance gaze clarity, resulting in super-resolved images $I^{\text{SR}}$.
		
		\STATE \textbf{Step 7: Apply ShanghaiTechGaze Model for Gaze Prediction}
		\STATE Given face image $\mathbf{f}$, use 3D reconstruction to generate multiple perspectives of eye appearance.
		\STATE Update gaze prediction model $\mathcal{M}_{\text{Gaze}}$ using retrained data from expanded eye perspectives, obtaining the gaze coordinate $\mathbf{g}_{pred}$.
		
		\STATE \textbf{Step 8: Coordinate Consistency in Spatial Alignment}
		\STATE Remap ShanghaiTechGaze coordinates $\mathbf{g}_{pred}$ to 360° panoramic coordinates, ensuring that both $\mathbf{g}_{pred}$ and $\Pi_{\alpha}^{2D}$ share the same spatial reference.
		
		\STATE \textbf{Step 9: Object Detection for Gaze Accuracy Enhancement}
		\STATE Use object detection to identify target object $\mathcal{O}$ near $\mathbf{g}_{pred}$.
		\STATE Define gaze precision:
		\[
		\epsilon = \begin{cases} 
			\text{high} & \text{if } \mathbf{g}_{pred} \in \mathcal{O} \\ 
			\text{low} & \text{if } \mathbf{g}_{pred} \notin \mathcal{O}
		\end{cases}
		\]
		\STATE Evaluate prediction accuracy based on $\epsilon$.
		
		\STATE \textbf{Step 10: Evaluation of Prediction Accuracy}
		\STATE Define accuracy criterion: If $\mathbf{g}_{pred}$ lies within $\mathcal{O}$, it is considered accurate; otherwise, it is inaccurate.
		
		\STATE \textbf{Step 11: Optimization through Iterative Testing}
		\FOR{each trial $t$ in testing phase}
		\STATE Adjust camera positions or parameters to maximize spatial coverage and precision.
		\STATE Log accuracy metrics $\epsilon_t$ and update $\mathcal{M}_{\text{Gaze}}$ as needed.
		\ENDFOR
		\STATE Return final optimized model $\mathcal{M}_{\text{Gaze}}^{*}$.
	\end{algorithmic}
	\label{algorithm2}
\end{algorithm}

\clearpage

\subsection*{EmoGaze360-1K and EmoGaze2D-50 datasets construction}
To create the EmoGaze360-1K dataset, we collected 1,000 panoramic images from platforms like 360cities and Flickr, encompassing a wide range of indoor and outdoor settings. When selecting these images, following PANDORA~\cite{Pandora}, we focused on three key criteria: 1) they depict real-world scenes, 2) they contain multiple instances of people and objects per image, and 3) they represent a diverse mix of indoor and outdoor environments. This approach ensures that the dataset aligns with real-world scenarios and applications. All images have a resolution of 1,920 × 960, providing high-quality visual information.

Unlike previous emotion recognition datasets that rely solely on visual stimuli (e.g., images and videos) to evoke emotions but don't incorporate them into their training or testing sets, EmoGaze360-1K includes both eye fixation data and contextual panoramic images. This dual-focus design allows for a deeper understanding of the relationship between visual attention and emotional states, providing a comprehensive dataset for emotion recognition research. Additionally, EmoGaze360-1K includes emotional annotations for six distinct emotional states across multiple modalities, including EEG signals, facial expressions, eye-tracking data, precise visual fixation points, and environmental context. This multimodal design offers more detailed and varied input data for more robust emotion recognition analysis.

To facilitate the collection of eye fixation data, we compiled a set of 500 emotion-inducing images and 50 emotion-inducing videos from open emotion database such as AffectNet and IAPS (International Affective Picture System) and Youtube, resulting in a total of 2,500 images and 250 videos for each emotional state. These resources are designed to evoke corresponding emotional responses from users before the eye fixation data collection stage.

Based on the panoramic fixation collection approach (i.e., WinDB~\cite{WinDB}, a head-mounted display (HMD)-free approach, which is more comfortable than the HMD-based one), we recruited 20 users, including 8 females and 12 males aged between 19–26. All users were completely unfamiliar with the fixation collection process, and none of the images from our pool had been shown to them previously. Note that with the WinDB approach, each user only needs to view the images (with a resolution of 1,920 × 960) on a PC, with a standard eye tracker set up to record the data. EEG and facial expression data were also collected in parallel during each session to capture additional emotional responses, ensuring that the multimodal annotations reflect both the eye-tracking and physiological aspects of emotion.

To ensure accurate fixation data collection, each user viewed 100 images corresponding to a single emotion in each session, with the entire fixation process lasting approximately 40 minutes. After a day's break, a second session was conducted to annotate a different emotion. To maintain emotional consistency, an emotional stimulus was administered every 20 images. The process could be paused at any time if the user experienced fatigue or discomfort. However, an emotional stimulus was applied before each annotation to ensure consistent emotional responses.

After collecting the eye fixation points, we generated scanpaths for each image using established Scanpath annotation methods~\cite{HAT,IndivScan}. These scanpaths were combined with EEG signals, facial expression data, and visual fixation points to create a rich, multimodal dataset that captures both cognitive and emotional responses to panoramic scenes. Note that the SEED-V-Multimodal dataset is also collected and annotated in this manner.

We now discuss the features of the proposed dataset and its advantages. It contains 1,000 panoramic images, including 800 indoor scenes and 200 outdoor scenes, spanning 52 categories. The dataset is divided into training and testing sets with a 70/30 ratio, allowing for robust ten-fold training. Here are the key advantages of EmoGaze360-1K:

1) Comprehensive Integration of Fixation Trajectories: Unlike existing datasets that focus on specific eye movement signals like pupil size or diameter, EmoGaze360-1K includes fixation trajectories that reflect interaction with the environment. This comprehensive approach allows for a deeper understanding of visual attention in relation to environmental context, resulting in more accurate emotion recognition. By combining these visual cues with EEG and facial expression data, EmoGaze360-1K provides an even richer understanding of how emotional states are expressed and perceived.

2) Non-Intrusive and Cost-Effective Data Collection: EmoGaze360-1K adopts an HMD-free approach, reducing the discomfort often associated with traditional methods like EEG and surface sensors. This design makes data collection more user-friendly and applicable in real-world scenarios where comfort and acceptance are crucial. Furthermore, using widely available eye-tracking equipment alongside EEG and facial expression recognition, this dataset minimizes the need for expensive, specialized hardware, making it more accessible for broader research applications.

The EmoGaze2D-50 dataset shares the same collection setup as EmoGaze360-1K, including its multimodal structure, emotional stimuli, and viewpoint tracking methodologies. Like EmoGaze360-1K, EmoGaze2D-50 incorporates data from multiple modalities, such as EEG signals, facial expressions, eye-tracking data, visual fixation points, and emotional annotations across six distinct emotional states. These data provide a rich foundation for studying the interplay between emotional responses and visual attention.
The primary difference lies in the type of visual content used for data collection. EmoGaze360-1K utilizes immersive 360-degree panoramic images viewed on flat-screen displays to simulate real-world environments, whereas EmoGaze2D-50 focuses on traditional 2D videos displayed on the same medium. 

\clearpage

\begin{figure*}[h]
	\centering
	\includegraphics[width=1\linewidth]{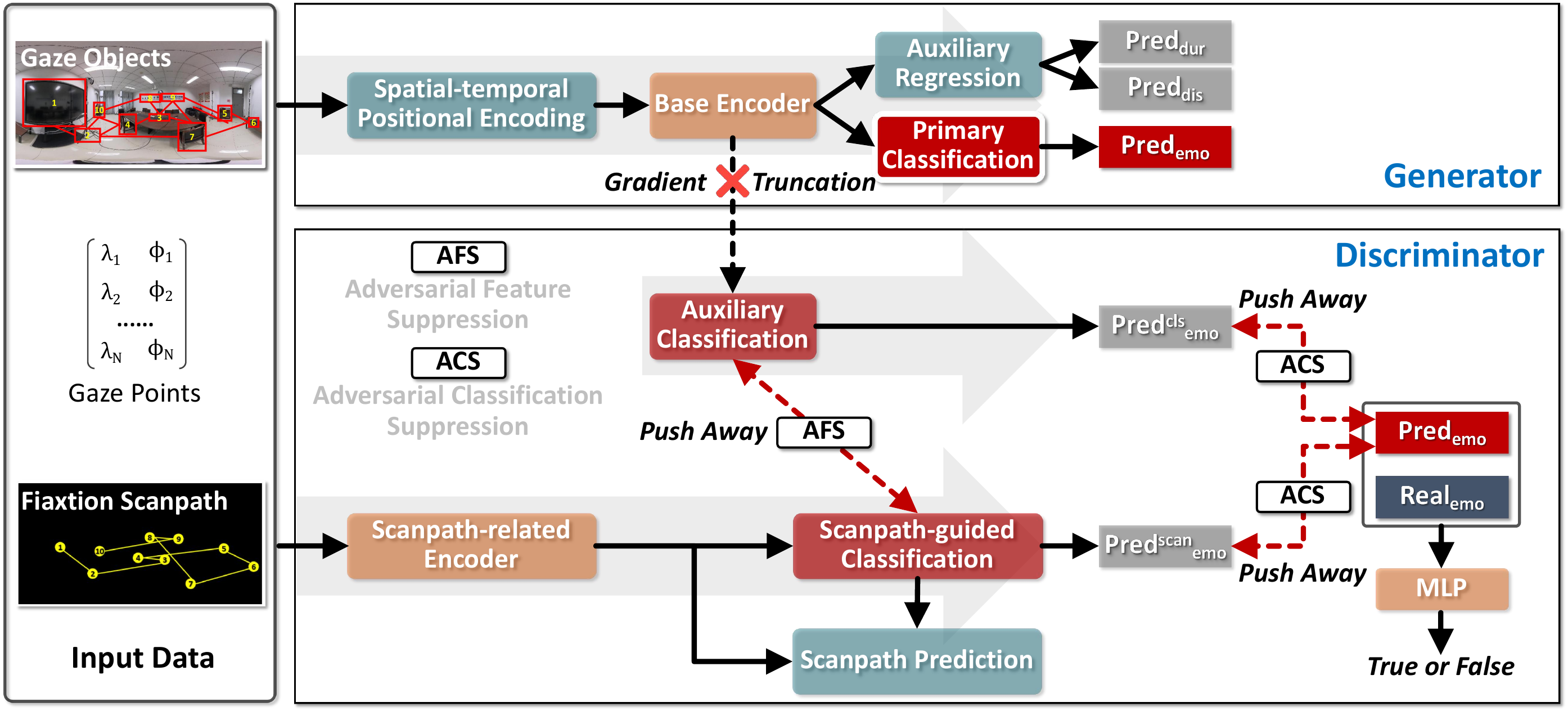}
	\vspace{-0.5cm}
	\caption{Pipeline of the proposed EmoGazeNet model. The proposed EmoGazeNet model consists of a Generator and a Discriminator. This approach leverages the varying gaze patterns towards different viewing object orders under different emotional states to enhance emotion recognition.}
	\label{fig:Pipeline}
	%\vspace{-0.3cm}
\end{figure*}

\subsection*{EmoGazeNet model architecture}
EmoGazeNet is an emotion recognition framework based on Generative Adversarial Networks (GANs). This model aims to accurately recognize different emotional states by integrating eye movement scanpath with environmental context. As shown in Supplementary Figure \ref{fig:Pipeline}, the design of EmoGazeNet consists of two main components: the Generator and the Discriminator. The Generator is responsible for generating the probability distribution of emotion categories, while the Discriminator's task is to distinguish between real and generated data. Through adversarial training, both the Generator and Discriminator improve over time, enhancing the accuracy and robustness of emotion recognition.

EmoGazeNet takes two primary inputs: (1) a panoramic ERP (Equirectangular Projection) image representing the environment, and (2) sequential gaze coordinates reflecting human-environment interactions. These gaze coordinates form a fiaxtion scanpath by sequentially connecting fixation points, which are then used to identify object regions within the ERP image. Specifically, the scanpath's fixation points determine which objects or regions in the environment received visual attention, and these regions are then extracted as separate patches. These patches are arranged according to the temporal sequence of the scanpath (i.e., the order in which they were viewed), creating what we call Semantic Interactive Orders (SIO). The key insight is that different emotional states produce distinctive viewing patterns, resulting in unique arrangements of environmental patches that serve as emotion signatures.

\subsubsection*{Generator of EmoGazeNet model} 
The Generator contains two modules: Spatial-temporal Positional Encoding, and Primary Classification. The input of the Generator is the detected sequencial object regions under six emotion states, instead of a random noise. The base encoder and primary classifier are combined together to predict the distribution of emotional states.

\textbf{Spatial-temporal Positional Encoding.} To calculate the position encoding for patch sequences under different emotional states, we propose Spatial-temporal Positional Encoding to incorporate temporal and spatial information from the eye movement scanpath into the position encoding.
The temporal information from the eye movement scanpath can be represented by the sequence of fixation times. For example, in the detected sequencial object regions under six emotion states, the first object corresponds to time $t_1$, the second object corresponds to time $t_2$, and so on. 
The spatial information of each object can be represented by its centroid position within the object regions.
Combine the temporal and spatial information to generate the position encoding. We can use a two-dimensional position encoding $(pos_{x,y})$ to represent the object's location within the image, and then add the temporal information to generate a three-dimensional position encoding $(pos_{x,y,t})$:
\begin{equation}
	pos_{x,y,t} = [pos_x,pos_y,t],
	\label{eq:PE}
\end{equation}
where $pos_x$ and $pos_y$ represent the object's position within the image. $t$ represents the patch's temporal order in the eye movement Scanpath.
Then, arrange the objects according to the viewing order under different emotional states. The detailed steps are as follows: generate patch embeddings for each object using a convolutional neural network, add the position encoding to the patch embeddings to obtain the patch embeddings with positional information. Input the patch embeddings with positional information into the Transformer encoder for emotion recognition.

\textbf{Primary Classification.} We consider the classification branch in the Generator as the Primary Classification Branch, whose output serves as the final emotion recognition results. The Primary Classification Branch consists of a Transformer encoder and a classification head. We adopt the Transformer encoder and MLP head of ViT~\cite{dosovitskiy2020image} as our Transformer encoder and classification head. Specifically, the Transformer encoder takes patch embeddings with positional information as input. 
As emotion recognition task can be considered as multi-classification problem, different from the MLP head in ViT, which directly predicts the category-specific labels, instead, our classification head transfer to predict the category probability distribution. 
In order to make the emotion recognition more accurate, we predicted both duration and dispersion as well. Note that this Auxiliary Regression Branch is independent of the generative adversarial network.

\subsubsection*{Discriminator of EmoGazeNet model}
Typically, after the Generator generates the emotion states probability distribution, a simple Discriminator is needed to distinguish the generated emotion states probability distribution and the real emotion states probability distribution. However, in this emotion recognition task, simply doing so poses the problem that the model may mechanically learn the ordering of different patches instead of performing deeper semantic learning, which will eventually lead to incorrect emotion recognition.
In emotion recognition tasks, simply relying on visual features may not be sufficient to accurately capture emotional states, as temporal and spatial information of viewpoint trajectories plays a key role in emotion recognition. To address this issue, we design an auxiliary classification branch and a scanpath-guided classification branch, and distancing the distance between ordinary classification features and scanpath-guided temporal and spatial features through an adversarial backward inhibition process to ensure that the adversarial learning process pays more attention to the temporal and spatial dependencies between the image patches in different emotional states.

\textbf{Auxiliary Classification.}
The auxiliary classification branch aims to enhance the discriminative power of the model by adding additional supervised signals to prevent model overfitting.
The architecture of auxiliary classification branch is the same as primary classification branch. The difference is that the auxiliary classification branch has an additional auxiliary classification neck to extract higher level features, thus enhancing the discriminative power of the features. The features output by the auxiliary classification neck access an auxiliary classification header, which maps the features to the category space through a number of fully connected layers. The last layer uses a Softmax activation function to output a probability distribution for each category.

\textbf{Scanpath-guided Classification.}
The Scanpath-guided classification branch aims to improve the accuracy of emotion recognition by capturing the temporal and spatial relationships between image patches by utilizing the coordinates of viewpoints in different emotional states to ensure that the model understands the contextual and sequential dependencies inherent in the data.
The Scanpath-guided classification branch consists of two parts: Scanpath-guided Classification and Scanpath Prediction. 

By introducing the Scanpath information, the model not only relies on the visual features of the image patches themselves, but also incorporates the information of the eye movement trajectories, making the emotion classification more accurate and reliable.
First, we use a Scanpath encoder (e.g., RNN or LSTM) to encode the sequence of viewpoint coordinates. The role of the Scanpath encoder is to form a time sequence representing the movement Scanpath that capture the temporal and spatial relationships of viewpoint movement in different emotional states.
The extracted Scanpath features are fed into a classification network, which consists of a neck (e.g., multiple convolutional) and a head (e.g., a fully-connected layer plus a Softmax activation function). The role of the neck is to further extract the high-level features, whereas the head maps the features to the category space and outputs a probability distribution for each category.

\textbf{Scanpath Prediction.}
The main role of Scanpath Prediction is to reconstruct the Scanpath of the gaze point. By adding the Scanpath prediction task, the model is able to learn both emotion recognition and Scanpath prediction tasks simultaneously. This multi-task learning approach ensures that the adversarial learning process focuses more on temporal and spatial dependencies between image patches in different emotional states, and enhances the generalization ability of the model.
The extracted Scanpath features by the Scanpath encoder are fed into the Scanpath prediction decoder to reconstruct gaze point trajectories.	
Scanpath reconstruction loss (e.g., mean square error loss) is used to measure the difference between the reconstructed Scanpath and the true Scanpath. By minimizing the Scanpath reconstruction loss, the model can gradually learn more accurate Scanpath patterns.
Joint training of the sentiment classification task and the Scanpath prediction task allows the two tasks to be mutually reinforcing by sharing some of the network layers, improving the overall model performance.

\textbf{Adversarial	Reverse Suppression.}
Emotion recognition requires not only considering the visual features of the image, but also understanding the temporal and spatial information of the viewpoint Scanpath, which is crucial for accurately capturing the emotional state. If the ordinary classification features and Scanpath-guided features are too similar, the model may not be able to effectively differentiate between visual features and temporal and spatial features, leading to insufficient understanding of emotion recognition. Through the adversarial backward inhibition process, the distance between the ordinary classification features and the Scanpath-guided features is distanced to avoid the two feature representations from being too similar, thus ensuring that the model pays more attention to the temporal and spatial dependencies between the image patches in different emotional states during the adversarial learning process.
To achieve, we propose Adversarial Reverse Suppression, which is implemented by an Adversarial Reverse Suppression loss, which includes Adversarial Feature Suppression and Adversarial Classification Suppression.  

For Adversarial Feature Suppression, it aims to pull out the features distance between the output features of Scanpath-guided Classification neck and Auxiliary Classification neck. This is achieved by Mutual information (MI), which is a measure used to quantify the dependency between two random variables. In deep neural networks, mutual information loss can be introduced to adjust the training objective of the network, thereby achieving the suppression or separation of specific information between features. Mutual information is defined as follows:
\begin{equation}
	I(X;Y)=\iint p(x,y)log\frac{p(x,y)}{p(x)p(y)} \mathrm{d}x \mathrm{d}y ,
	\label{eq:mi}
\end{equation}
where $X$ and $Y$ are two output features, $p(x,y)$ is the joint probability density function, and $p(x)$ and $p(y)$ are the marginal probability density functions of $X$ and $Y$, respectively. $I(X;Y)$ is the mutual information of $X$ and $Y$. Assuming we use approximation methods to estimate mutual information, the mutual information loss can be defined as:
\begin{equation}
	\mathcal{L}_{\text {mi }}=-I(X;Y),
	\label{eq:miloss}
\end{equation}

For Adversarial Classification Suppression, we first calculate the KL divergence between the category probability distributions from the Auxiliary Classification Branch and the Scanpath-Guided Classification Branch. The KL divergence measures the distance between two distributions, given by the formula:
\begin{equation}
	D_{KL}(P \| Q)=\sum_{i} P(i) \log \frac{P(i)}{Q(i)},
	\label{eq:kl}
\end{equation}
where $P$ and $Q$ represent the category probability distributions of the Auxiliary Classification Branch and the Scanpath-Guided Classification Branch, respectively.

Then, we design two suppression terms: adversarial suppression term and reverse suppression term. Add the KL divergence as an adversarial suppression term $Q$ to the loss of the Auxiliary Classification Branch to suppress the learning of Scanpath-guided features, while add the KL divergence $P$ as a reverse suppression term to the loss of the Scanpath-Guided Classification Branch to suppress the learning of common classification features.

The total loss of the Auxiliary Classification Branch is the sum of its classification loss ($\mathcal{L}_{\text {cls }}$, here we use Cross-Entropy Loss) and the adversarial suppression term $D_{K L}(P \| Q)$:
\begin{equation}
	\mathcal{L}_{\text {aux\_cls }}=\mathcal{L}_{\text {cls }}+\lambda D_{K L}(P \| Q),
	\label{eq:Auxiliary}
\end{equation}
where $\lambda$ is a balancing parameter that adjusts the weight between the classification loss and the adversarial suppression term.

The total loss of the Scanpath-Guided Classification Branch is the sum of its classification loss $\mathcal{L}_{\text {cls }}$, Scanpath reconstruction loss $L_{\text {rec }}$, and the reverse inhibition term $D_{K L}(Q \| P)$:
\begin{equation}
	\mathcal{L}_{\text {traj }}=\mathcal{L}_{\text {cls }}+L_{\text {rec }}+\beta D_{K L}(Q \| P),
	\label{eq:Scanpath}
\end{equation}
where $\beta$ is a balancing parameter that adjusts the weight between the classification loss, Scanpath reconstruction loss, and the reverse suppression term.

Combine the total losses of the Auxiliary Classification Branch and the Scanpath-Guided Classification Branch to form the Adversarial	Reverse Suppression loss:
\begin{equation}
	\mathcal{L}_{\rm ars\_total}=\mathcal{L}_{\rm traj}+\mathcal{L}_{\rm aux\_cls}+\mathcal{L}_{\text {mi }},
	\label{eq:adv_rev_sup}
\end{equation}

\subsubsection*{Overall Loss Function}
\label{sec:loss}
In addition to the network structure, the design of suitable loss functions is also essential for deep learning models. 
In this work, the loss functions are divided into two parts: Generator-related loss and Discriminator-related loss.

We propose two different loss functions for the Generator, including the regression loss ($\mathcal{L}_{reg}$), and the minmax loss.

Unlike the common deep learning models, the GAN framework is optimized by an adversarial training procedure. The effect of generating real data is achieved by optimizing the adversarial relationship between the Generator and the Discriminator using the minmax loss $\mathcal{L}_{adv}$.

The total loss function of the Generator is the weighted sum of each of the above losses:
\begin{equation}
	\begin{array}{c}
		\mathcal{L}_{\rm G\_total}=\mathcal{L}_{reg}+\mathcal{L}_{adv}.
	\end{array}
\end{equation}

We propose two different loss functions for the Discriminator, including the adversarial suppression loss ($\mathcal{L}_{\rm ars\_total}$, Supplementary Equation 7) and the categorical cross-entropy loss ($\mathcal{L}_{\rm cat\_ce}$).

To ensure accurately reconstructing the viewpoint Scanpath, we use a mean squared error Loss (MSE) and dynamic time warping (DTW). Specifically, DTW is used to measure the similarity between two time series (e.g., trajectories) that can handle nonlinear deformations on the time axis, MSE is used to directly measure the error between reconstructed and true Scanpath points.
Assuming that the scanpath $x=[x_1,x_2,...,x_n]$ is the true scanpath, $\hat{x}=[\hat{x}_1,\hat{x}_2,...,\hat{x}_n]$ is the reconstructed scanpath and the total Scanpath reconstruction loss function can be written in the following form:
\begin{equation}
	\begin{array}{c}
		\mathcal{L}_{\text {rec}} = \alpha\times  \mathcal{L}_{\text {mse}}(\hat{x}_i,x_i)+\beta\times  \mathcal{L}_{\text {dtw}}(\hat{x}_i,x_i),
	\end{array}
\end{equation}
where $\hat{x}_{i}$ is the reconstructed scanpath point and $x_{i}$ is the true Scanpath point. $\alpha$ and $\beta$ are weighting coefficients to balance the effects of the two components of the loss. The mean squared error Loss can be formulated by:
\begin{equation}
	\begin{array}{c}
		\mathcal{L}_{\text {mse}} = \frac{1}{n} \sum_{i=1}^{N}\left(\hat{x}_{i}-x_{i}\right)^{2},
	\end{array}
\end{equation}

Specifically, DTW looks for a path $P=[p_1,p_2,...,p_L]$ where $p_l=(i_l,j_l)$ means that the $i_l$-th point in $x$ matches the $j_l$-jth point in $\hat{x}$ such that the total distance:
\begin{equation}
	\begin{array}{c}
		\mathcal{L}_{\text {dtw}} = \underset{P}{\rm min}\sum_{l=1}^{L}d(\hat{x}_{jl},x_{il}) ,
	\end{array}
\end{equation}
where $d(\hat{x}_{jl},x_{il})$ is the distance between the matching points in $x$ and $\hat{x}$, which is usually used as the Euclidean distance.

In this work, the Discriminator is trained to distinguish between real and fake data by using a categorical cross-entropy loss. In this case, the output layer of the discriminator will usually be a softmax layer that outputs a probability distribution for each category. The loss function can be defined as:
\begin{equation}
	\mathcal{L}_{\rm cat\_ce}=-\sum_{i=1}^{N}\sum_{c=1}^{C} y_{ic} \log \left(p_{ic}\right),
\end{equation}
where $N$ is the number of samples, $C$ is the number of categories, $ y_{ic}$ is the true label of sample $i$ for class $c$, and $p_{ic}$ is the predicted probability that sample $i$ belongs to class $c$.

The total loss function of the Discriminator is the weighted sum of each of the above losses:
\begin{equation}
	\begin{array}{c}
		\mathcal{L}_{\rm D\_total}=\mathcal{L}_{\rm ars\_total}+\mathcal{L}_{\rm cat\_ce}+\mathcal{L}_{\text {dtw}}+\mathcal{L}_{\text {mse}}+\mathcal{L}_{\text {rec}}.
	\end{array}
\end{equation}

The total loss function of the Generator and the Discriminator is as follows:
\begin{equation}
	\begin{array}{c}
		\mathcal{L}_{\rm total}=\mathcal{L}_{\rm D\_total}+\mathcal{L}_{\rm G\_total}.
	\end{array}
\end{equation}

\subsection*{Detailed explanations of six indicators in the experiment of emotion-environment interactions across various gender and personality}

(1) Emotion Sensitivity: This measures the system’s responsiveness to slight emotional changes in users. Higher scores indicate greater sensitivity, making it especially suited for capturing the frequent emotional shifts seen in extroverts and females.

(2) Emotion Stability: This reflects the continuity and consistency of emotional states. Higher scores indicate greater stability in similar environments, as typically seen in males and introverts, who tend to exhibit steadier emotional responses.

(3) Real-Time Response Capture: This measures the system’s speed in detecting rapid emotional changes. Higher scores indicate quicker responses, especially effective for capturing emotional shifts in dynamic situations, often seen in extroverts and females.

(4) Emotion Saliency Focus: This indicates the system’s focus on emotionally significant targets (such as highly emotional objects). Higher scores mean the system better identifies emotionally salient objects, with extroverts and females often displaying stronger responsiveness in this area.

(5) Context Adaptability: This measures the system’s adaptability across different environmental contexts. Higher scores indicate stronger adaptability, making it well-suited for extroverts and females who, due to their emotional sensitivity, are more compatible with such adaptive system features.

(6) Sustained Attention Preference: This assesses the system’s sensitivity to users’ preference for prolonged focus on emotionally salient objects. Higher scores indicate that the system can more accurately capture sustained attention behaviors, with males and introverts often displaying longer attention spans on specific objects.

\newpage

\section*{Supplementary Results}
\subsection*{Quantitative component studies of EmoGazeNet model}
We evaluated each component of the proposed emotion recognition model. As shown in Supplementary Table \ref{tab:component}, the evaluation matrix clearly demonstrates that the system reaches its peak performance in terms of ACC (80.22), F1 Score (78.86), and $caw{\rm F1}$ score (72.22) when all key components, such as STPE and PC, are engaged, highlighting the synergistic impact of these elements on overall system performance. Omitting certain components, like STPE, leads to a notable decrease in system performance, indicating that they play an essential role in the emotion recognition process. Components such as AR and ACH, when incorporated, can moderately enhance the system's accuracy and F1 score, yet their contribution is less pronounced compared to the core components like PC and BEn. Varied combinations of components result in significant performance fluctuations. For instance, the contrast between row 8 and row 11 illustrates that employing all adversarial suppression modules (AFS, ACS) results in superior system performance compared to most other configurations, underscoring the significance of adversarial suppression in curbing overfitting and bolstering the model's generalizability. The multiplicity of these branches fortifies the system's predictive capabilities, particularly within intricate emotion recognition tasks.

\begin{table}[h]
	\centering
	\caption{Quantitative evidence of component studies of EmoGazeNet model. \textbf{``$\uparrow$'': the higher values the better.} }
	%\vspace{-0.2cm}
	\begin{tabular}{c}
		%\begin{minipage}{1.01\linewidth}
		\hspace{-0.35cm}
		\includegraphics[width=1\linewidth]{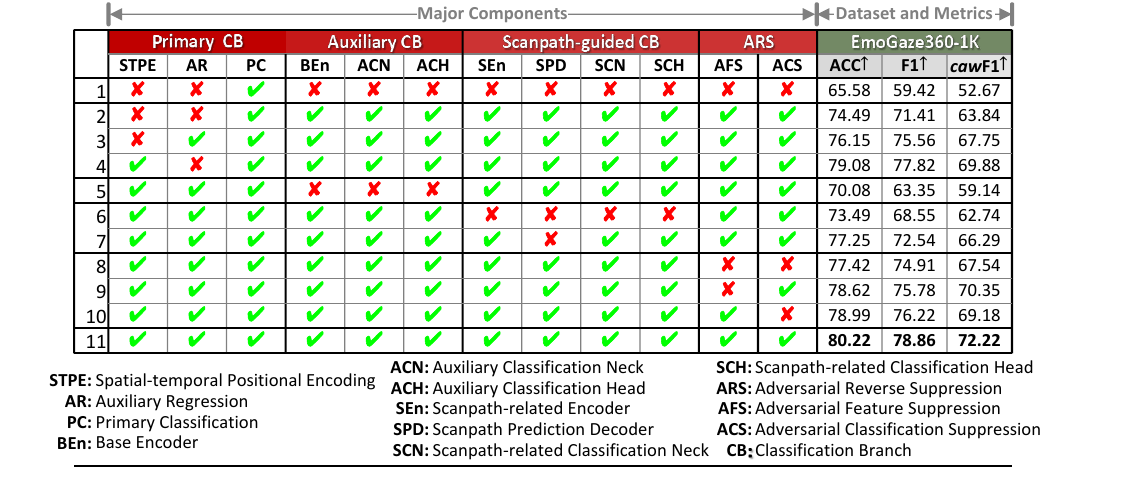}
		%\end{minipage}
	\end{tabular}
	\label{tab:component}
	\vspace{-0.2cm}
\end{table}

\clearpage

\subsection*{Different choices of scanpath prediction methods and base encoders in the Generator of EmoGazeNet model}
\begin{figure*}[h]
	\centering
	%	\hspace{-3.7cm}
	\includegraphics[width=0.8\linewidth]{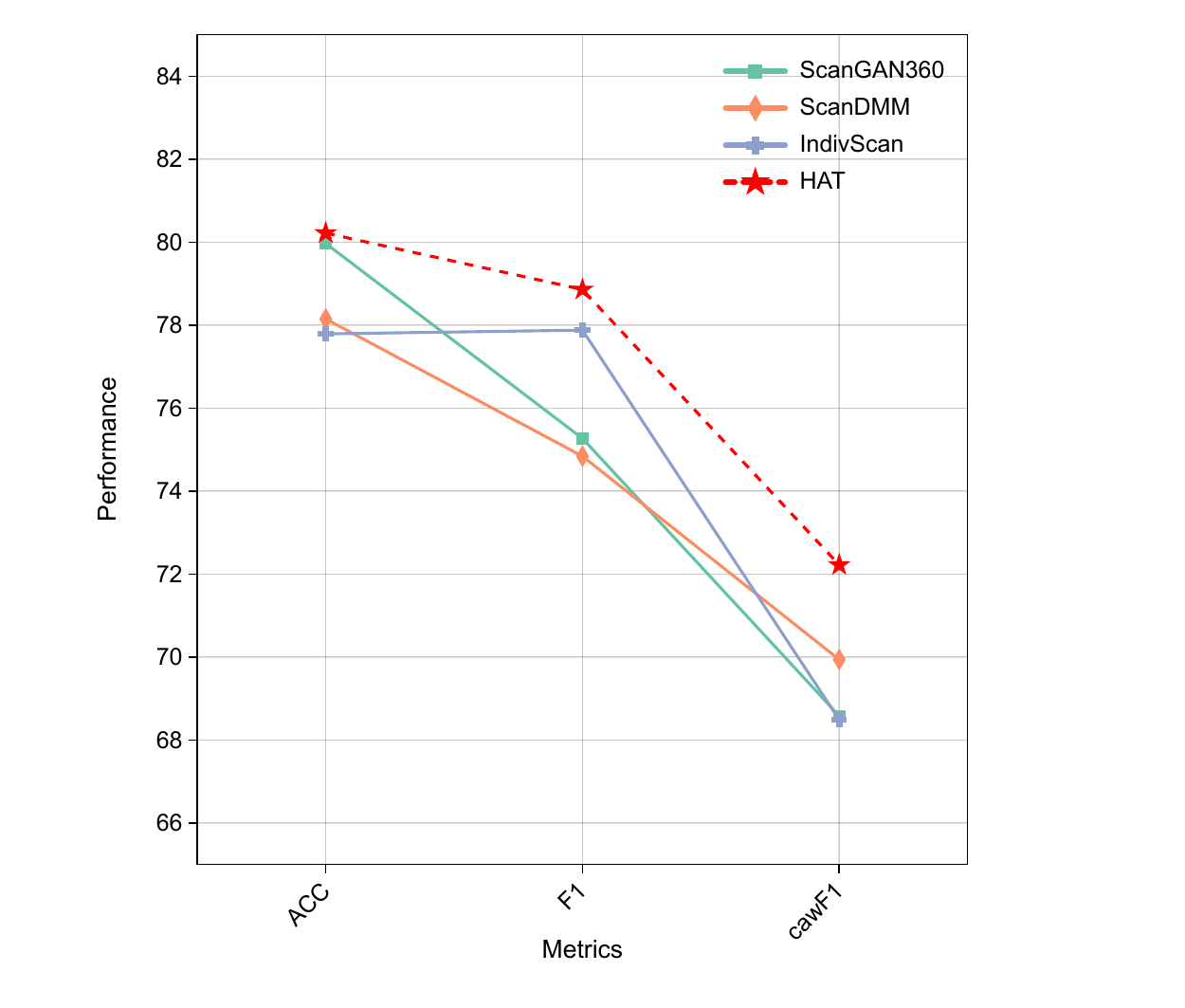}
	\vspace{-0.8cm}
	\caption{Performance comparison of different scanpath prediction methods (HAT~\cite{HAT}, IndivScan~\cite{IndivScan}, ScanGan360~\cite{ScanGan360}, and ScanDMM~\cite{ScanDMM}) in the Generator of EmoGazeNet model. }
	\label{tab:ScanPath}
	\vspace{-0.3cm}
\end{figure*}

\begin{figure*}[h]
	\centering
	%	\hspace{-3.7cm}
	\includegraphics[width=0.8\linewidth]{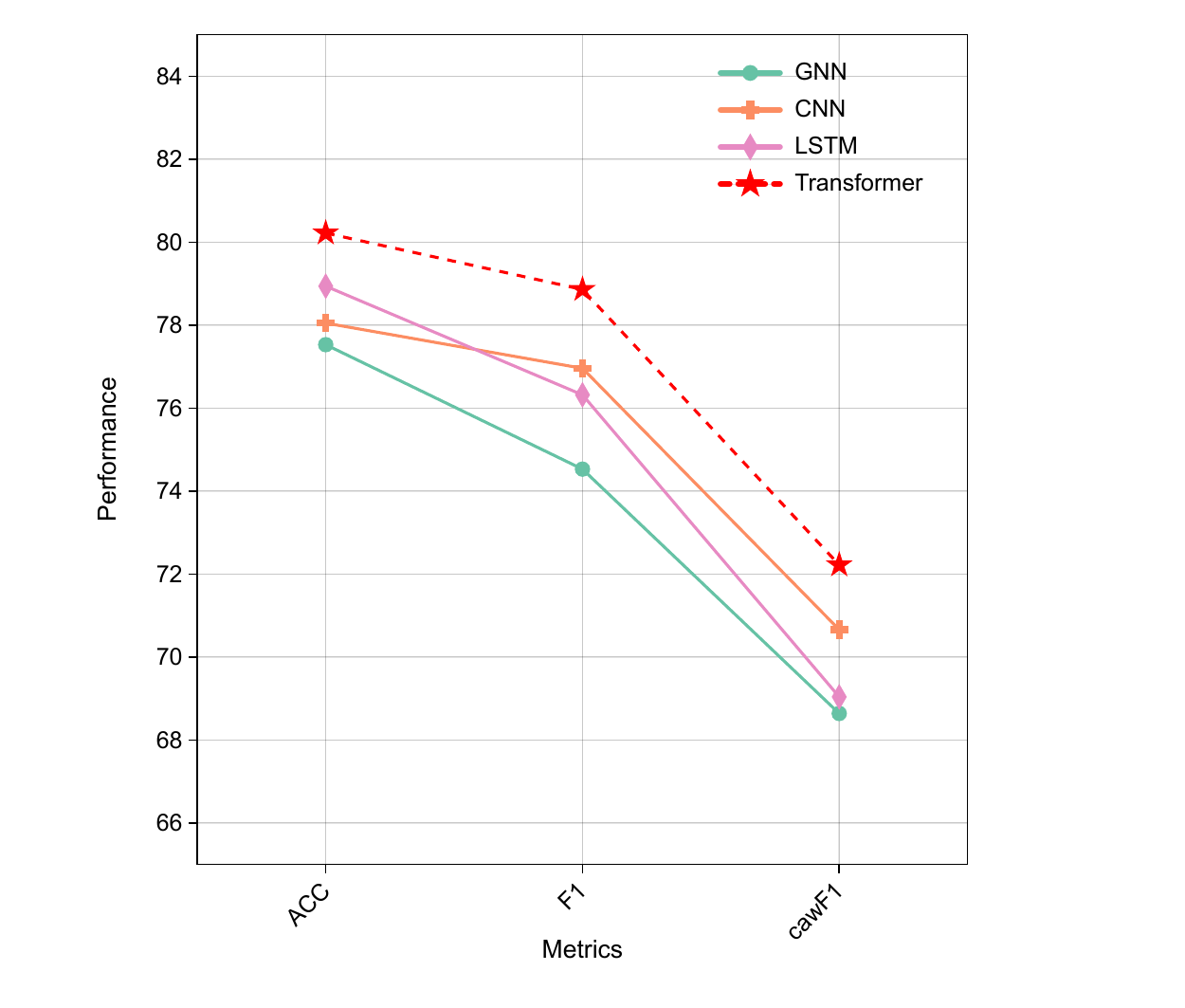}
	\vspace{-0.8cm}
	\caption{Performance comparison of different choices of base encoders (GNN~\cite{GNN}, CNN~\cite{CNN}, LSTM~\cite{LSTM}, and Transformer~\cite{Transformer}) in the Generator of EmoGazeNet model. }
	\label{tab:BaseEncoder}
	\vspace{-0.3cm}
\end{figure*}

We also compared different scanpath prediction methods and different choices of base encoder in the Generator (Supplementary Figure \ref{tab:ScanPath} \& \ref{tab:BaseEncoder}). Our findings indicate that employing HAT for scanpath prediction significantly enhances the $caw{\rm F1}$ score of emotion recognition, and leveraging a Transformer as the base encoder yields superior performance. HAT leverages a hierarchical attention mechanism, allowing it to more accurately capture the temporal and spatial features within eye movement data. Subtle changes in scanpaths are crucial for emotion recognition, and HAT can dynamically focus on these key points and regions, ensuring that critical information is preserved and enhanced during model processing. This mechanism results in more precise scanpath predictions, thereby improving the overall $caw{\rm F1}$ score of emotion recognition. The Transformer model has demonstrated strong capabilities in handling sequential data, such as scanpaths, particularly due to its self-attention mechanism, which processes the entire sequence in parallel rather than step-by-step like traditional RNNs or CNNs. This parallel processing not only enhances computational efficiency but also allows the model to better capture long-range dependencies and complex emotional patterns. Furthermore, the flexibility and scalability of the Transformer enable it to adapt more effectively to various emotional features, resulting in superior performance across different emotional states.

\clearpage

\subsection*{Different choices of gaze point prediction methods in eye appearance acquisition and gaze point generation process}
\begin{figure*}[h]
	\centering
	%	\hspace{-3.7cm}
	\includegraphics[width=0.8\linewidth]{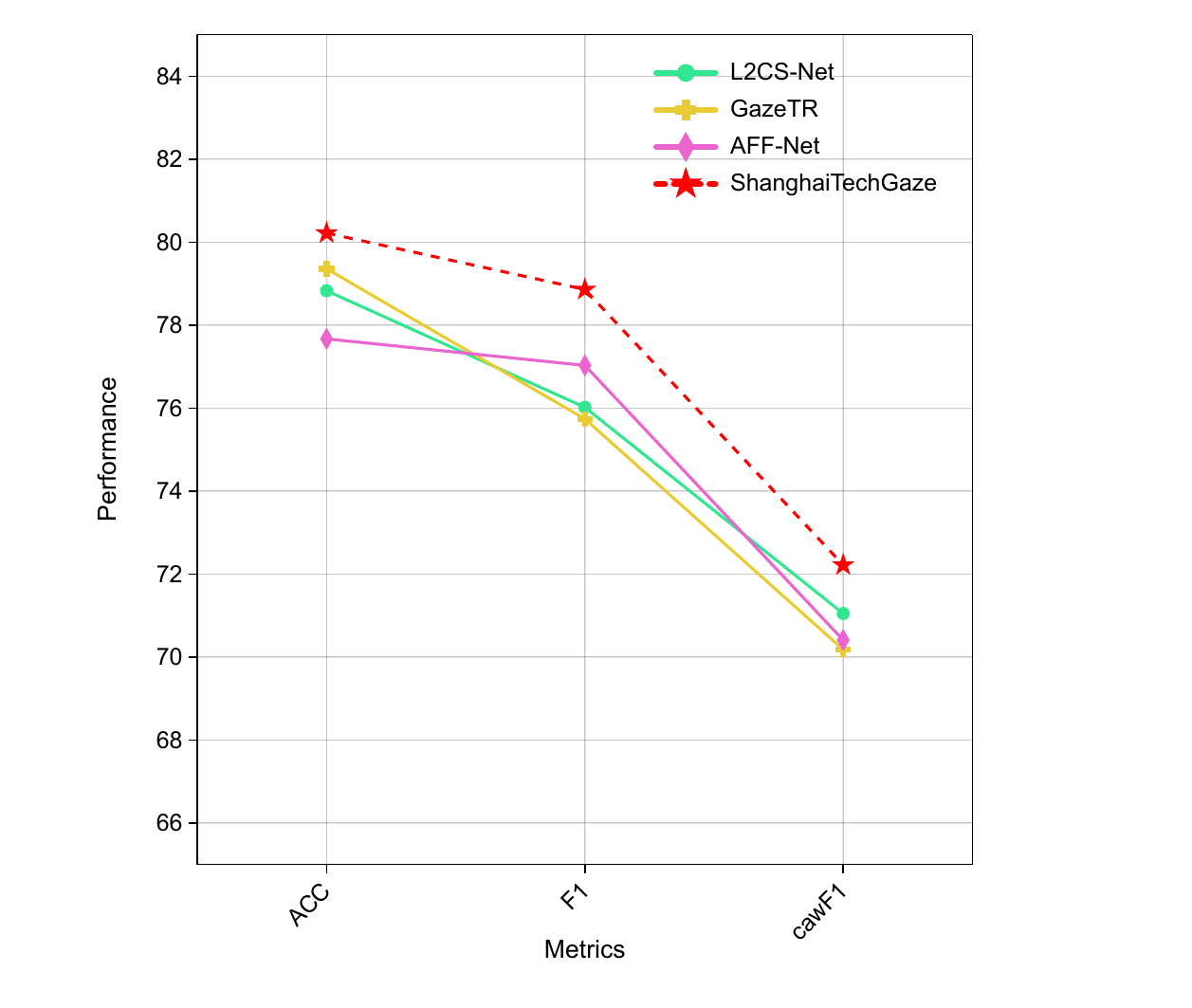}
	\vspace{-0.8cm}
	\caption{Performance comparison of different gaze point prediction methods (ShanghaiTechGaze~\cite{ShanghaiTechGaze}, L2CS-Net~\cite{L2CS-Net}, GazeTR~\cite{GazeTR}, and AFF-Net~\cite{AFF-Net}.) }
	\label{tab:GazePointPre}
	\vspace{-0.3cm}
\end{figure*}

Supplementary Figure \ref{tab:GazePointPre} shows a performance comparison across various gaze point prediction methods: ShanghaiTechGaze, L2CS-Net, GazeTR, and AFF-Net. ShanghaiTechGaze consistently outperforms the other methods across all metrics (ACC, F1, and $caw{\rm F1}$), indicating its robustness in gaze point prediction tasks. This could be attributed to the relatively small domain shift between the training and testing datasets. The data presented in these scenarios might be similar in format and content, and the way it is mapped onto the screen and subsequently captured is consistent. This reduced domain shift ensures that the model can generalize well from the training data to new instances, thereby enhancing its accuracy and overall performance in gaze point prediction tasks.

\clearpage

\subsection*{Different projection intervals during eye appearance acquisition and gaze point generation}
\begin{figure*}[h]
	\centering
	%	\hspace{-3.7cm}
	\includegraphics[width=0.8\linewidth]{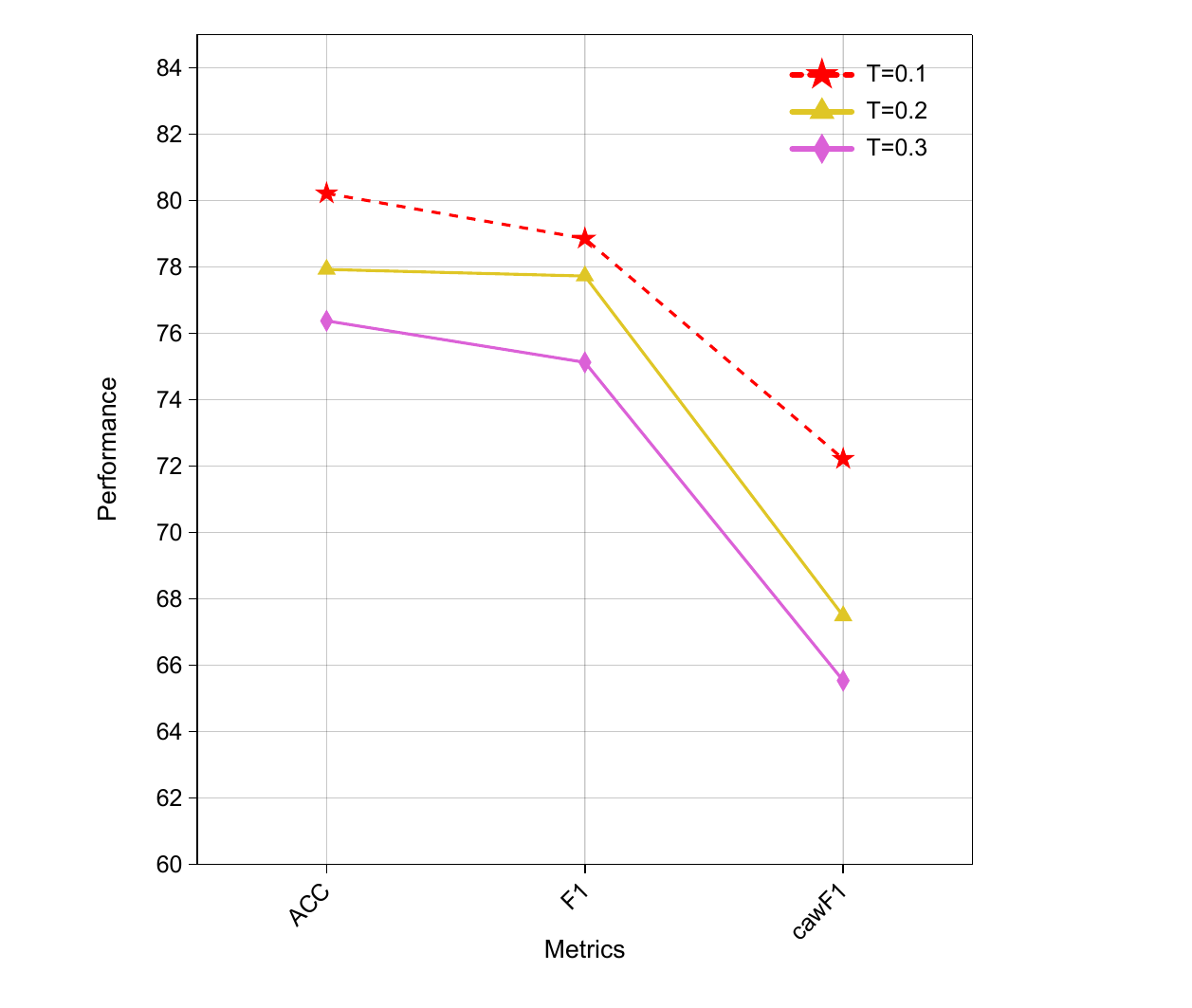}
	\vspace{-0.8cm}
	\caption{Performance comparison of different projection intervals (T=0.1, T=0.2, T=0.3) during eye appearance acquisition and gaze point generation. }
	\label{tab:ProjInterval}
	\vspace{-0.3cm}
\end{figure*}

Supplementary Figure \ref{tab:ProjInterval} shows the performance comparison of different projection intervals (T=0.1, T=0.2, T=0.3) across three metrics: Accuracy (ACC), F1 score, and $caw{\rm F1}$ score.
T=0.1 (Red star, dashed line) consistently outperforms the other projection intervals across all metrics, which can be attributed to the finer granularity of projection, allowing the model to capture more detailed and subtle changes in the gaze patterns. This finer interval helps in preserving crucial temporal information that is essential for accurate gaze prediction and, consequently, emotion recognition. By closely following the eye movement data, the model is less likely to miss significant transitions or small yet important shifts in gaze, leading to better overall performance.

% 数据采集方法  EmoGaze360数据量  不同摄像头数量  投影时间间隔

\clearpage
%\subsection*{Different number of HD cameras during eye appearance acquisition and different number of EmoGaze360-1K dataset in training EmoGazeNet model}
%Supplementary Figure \ref{tab:NumOfCam} shows the performance variations when using different number of HD cameras during eye appearance acquisition. The use of 8 cameras appears to strike an optimal balance between capturing comprehensive eye appearance data and managing the computational complexity and potential data redundancy. Eight cameras provide sufficient coverage of the eye region to capture the necessary details for accurate gaze prediction without overwhelming the system with excessive data. This balance helps maintain high accuracy and F1 scores, as the model can efficiently process the captured data without being hindered by unnecessary information. 

\subsection*{Different number of EmoGaze360-1K dataset in training EmoGazeNet model}
\begin{figure*}[h]
	\centering
	%	\hspace{-3.7cm}
	\includegraphics[width=0.8\linewidth]{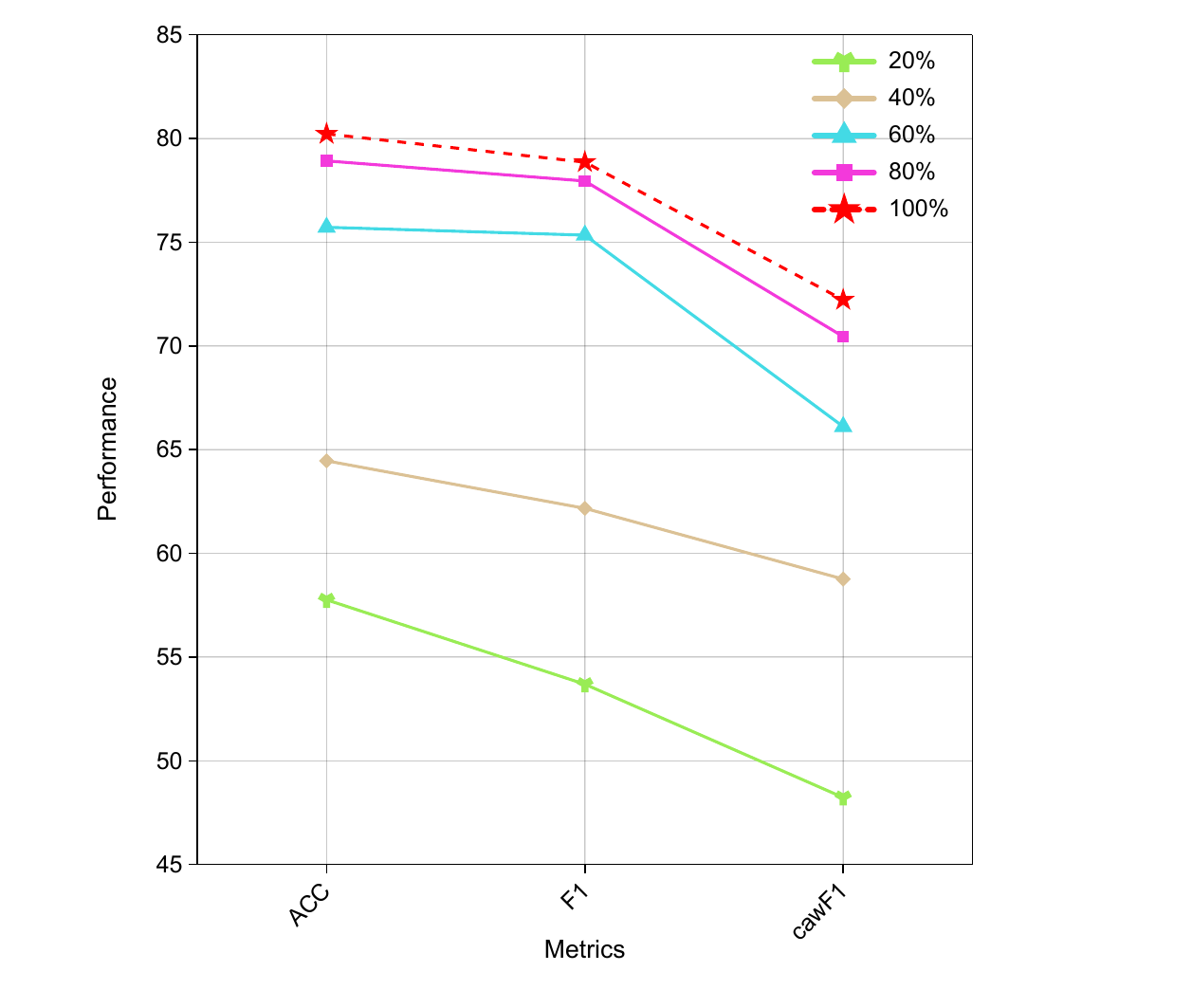}
	\vspace{-0.8cm}
	\caption{Performance comparison of different number of EmoGaze360-1K dataset in training EmoGazeNet model. }
	\label{tab:NumOfEmoGaze360}
	\vspace{-0.3cm}
\end{figure*}

Supplementary Figure \ref{tab:NumOfEmoGaze360} shows the performance variations when different proportions (20\%, 40\%, 60\%, 80\%, and 100\%) of the EmoGaze360-1K dataset are used to train the EmoGazeNet model. Our findings indicate that 100\% of the Dataset (Red star, dashed line) shows the highest performance across all metrics, particularly excelling in ACC and F1 scores. 
Using the entire EmoGaze360-1K dataset provides the model with the most comprehensive and diverse set of training examples. This extensive dataset allows the model to learn a wide range of features and patterns, leading to superior generalization and performance across all metrics. The full dataset likely covers more variations in gaze patterns, facial expressions, and other factors crucial for accurate emotion recognition, thus enhancing the model's ability to perform well in diverse scenarios.  
As the dataset size decreases, the model is exposed to fewer examples during training, which limits its ability to learn the full range of features present in the EmoGaze360-1K dataset. This reduction in data leads to a narrower understanding of the possible variations in gaze patterns, which is reflected in the declining performance metrics. The model may start to overfit to the smaller dataset, where it learns specific details rather than generalizable patterns, resulting in poorer performance, particularly in metrics like $caw{\rm F1}$ score, which are sensitive to class-wise variations.
The performance of the model trained on 80\% of the dataset is relatively close to that of the full dataset because 80\% might still cover most of the critical variations necessary for the model to generalize well. However, as seen in the slight drop in $caw{\rm F1}$ score, some finer details or rare cases that are present in the full dataset might be missing, leading to slightly reduced performance in handling specific classes or conditions.

\clearpage

\subsection*{Scanpath visualization of real 360-degree scenes}
For real 360-degree static image scenes, as illustrated in Supplementary Figure \ref{fig:visReal1}-A, under positive emotions, males primarily focus on prominent objects such as computer screens or slogans on the wall, then scan the workspace, and finally notice the light from outside the window. Females start with details on the desk, such as clutter or office supplies, then gradually observe the layout of the entire room, and finally pay attention to the slogans on the wall and the light from outside the window. Under negative emotions, males' gazes concentrate on the dark or cluttered areas on the floor and desk, paying less attention to the bright parts. Females start from the corners or cluttered areas of the room, then notice the details on the desk, but their gaze lingers more in the dark parts of the room. 

For dynamic scenes, such as two people moving along a fixed route in this corridor, as illustrated in Supplementary Figure \ref{fig:visReal1}-B, under positive emotions, males prioritize focusing on the moving figures, then turn to observe the overall layout and lighting of the corridor. Females start observing details of the figures (such as gait, attire), then focus on the posters on the wall and the scenery outside the window, and finally examine the layout of the entire scene. Under negative emotions, males concentrate their attention on the corners or inconspicuous places of the corridor, paying less attention to the dynamic figures. Females start from the shadowy or darker areas, then notice the moving figures, but their gaze may be brief. 

In high-light scenes, under positive emotions, as illustrated in Supplementary Figure \ref{fig:visReal2}-A, males focus on prominent objects on the display screen or desk, then scan the main areas of the meeting room, such as seats and slogans. Females start observing details on the desk, such as wires or documents, then gradually expand to the entire room's layout, including slogans on the wall and the window. Under negative emotions, males mainly focus on the dark parts of the meeting room or clutter on the floor, paying less attention to the bright areas. Females start from the corners or cluttered areas of the room, then notice the slogans on the wall, but their gaze lingers more on the untidy parts. 

In low-light scenes, as illustrated in Supplementary Figure \ref{fig:visReal2}-B, under positive emotions, males' gazes concentrate on the brighter areas, such as the window, then quickly scan the conference table and display screen. Females start with the well-lit window, gradually observe the details on the desk, and finally pay attention to the walls and slogans. Under negative emotions, males mainly focus on the shadows and darker areas of the room, paying less attention to the bright parts. Females start from the shadowy areas or clutter on the desk, then notice the window, but their gaze lingers more in the dark areas.

%\FloatBarrier

%%%%%%%%%%%%%%%%%%%%%%%%%%%%%%%%%%%%%%%%%%

\begin{figure*}[h]
	\centering
	\includegraphics[width=1\linewidth]{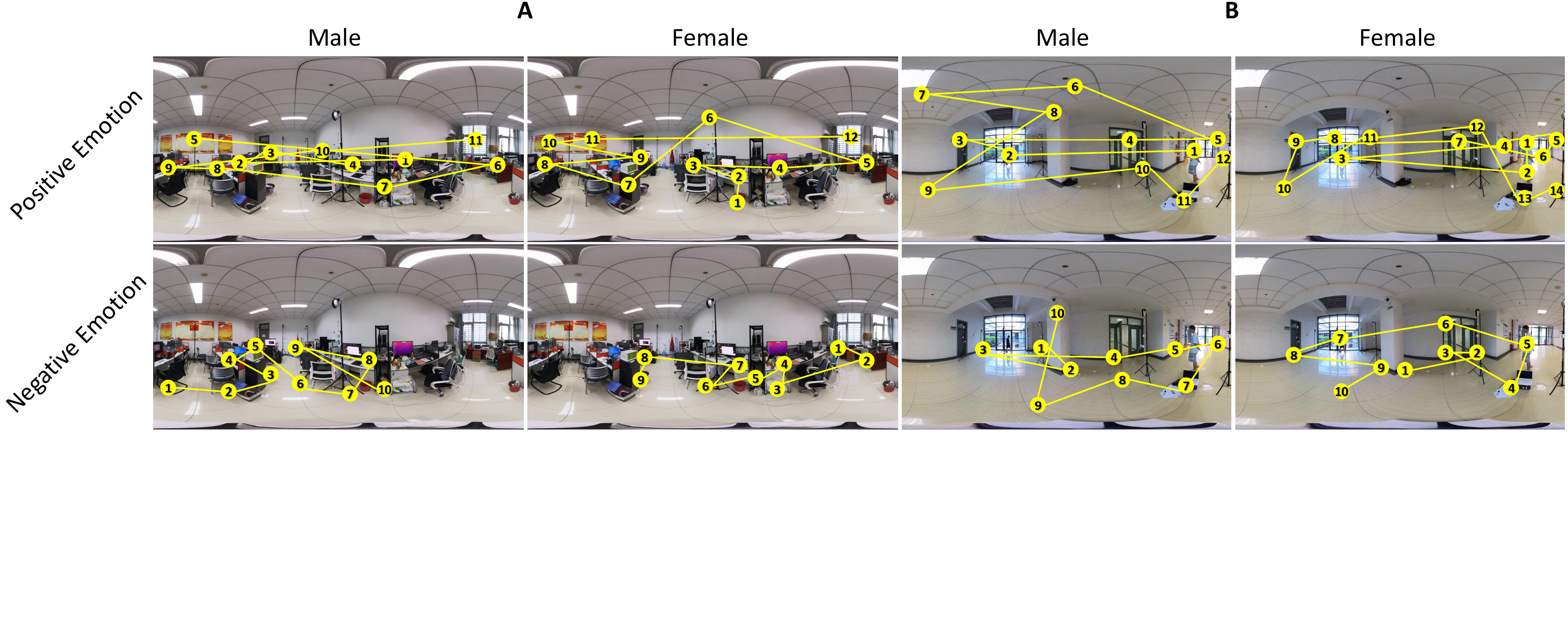}
	\vspace{-2.4cm}
	\caption{Scanpath visualization of real 360-degree static (A) and dynamic (B) scenes in different genders and emotion states.	``Positive Emotion'': happy; ``Negative Emotion'': fear, sad, disgust and angry.}
	\label{fig:visReal1}
	\vspace{-0.3cm}
\end{figure*}

\begin{figure*}[h]
	\centering
	\includegraphics[width=1\linewidth]{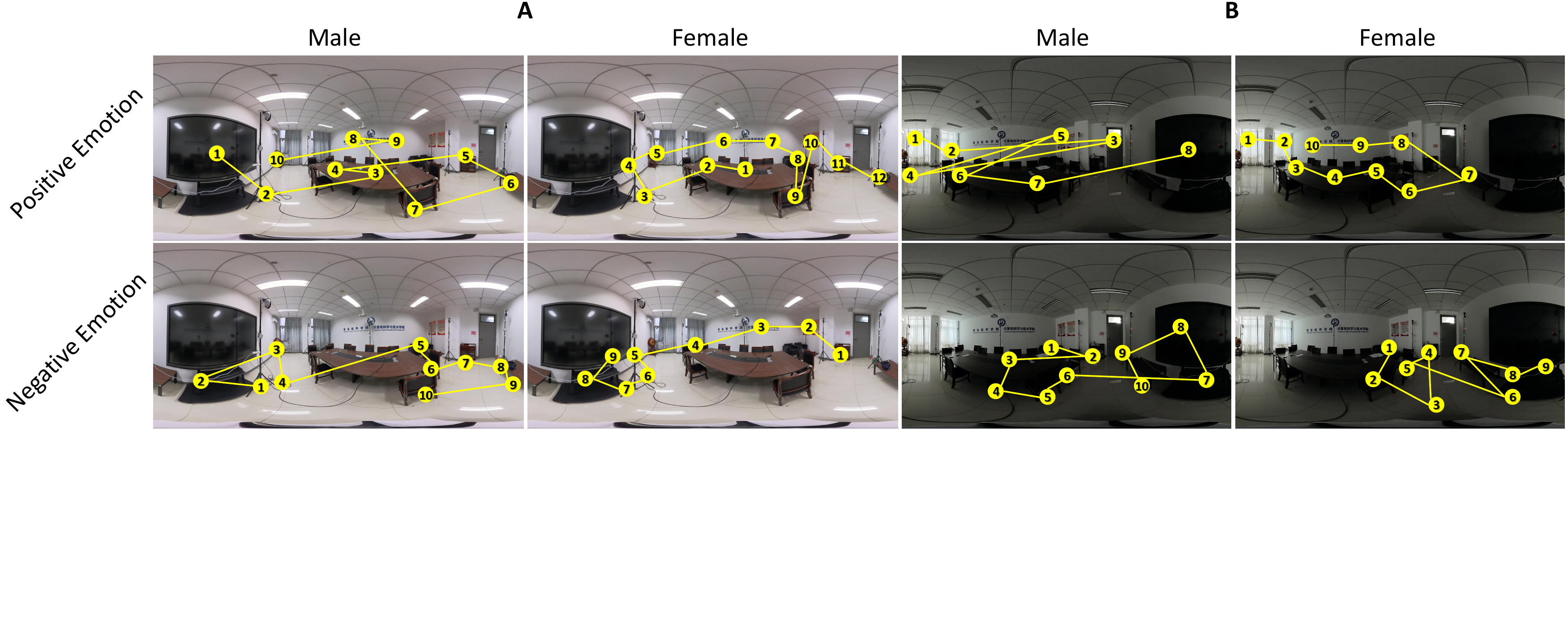}
	\vspace{-2.4cm}
	\caption{Scanpath visualization of real 360-degree high-light (A) and low-light (B) scenes in different genders and emotion states.	``Positive Emotion'': happy; ``Negative Emotion'': fear, sad, disgust and angry.}
	\label{fig:visReal2}
	\vspace{-0.3cm}
\end{figure*}

\end{document}